\documentclass{article}
\usepackage{microtype}
\usepackage{graphicx}
\usepackage{subfigure}
\usepackage{booktabs} 
\usepackage{csvsimple}

\usepackage[accepted]{icml2025}

\def\matern{Mat\'ern\,}
\usepackage{amsmath}
\usepackage{amssymb}
\usepackage{mathtools}
\usepackage{amsthm}
\usepackage{natbib}
\usepackage{xfrac}

\def\cite{\citep}

\theoremstyle{plain}

\usepackage[textsize=tiny]{todonotes}

\icmltitlerunning{Feature maps for the Laplacian kernel and its generalizations}

\usepackage{parthe_macros}

\begin{document}

\twocolumn[
\icmltitle{Feature maps for the Laplacian kernel and its generalizations}

\begin{icmlauthorlist}
\icmlauthor{Sudhendu Ahir}{cminds}
\icmlauthor{Parthe Pandit}{cminds}
\end{icmlauthorlist}

\icmlaffiliation{cminds}{Center for Machine Intelligence and Data Science, IIT Bombay, India}

\icmlcorrespondingauthor{}{pandit@iitb.ac.in}

\icmlkeywords{Kernel machines, Laplacian kernel, Mattern kernel, Exponential power kernel, Random Fourier features}

\vskip 0.3in
]

\printAffiliationsAndNotice{} 

\begin{abstract}
Recent applications of kernel methods in machine learning have seen a renewed interest in the Laplacian kernel, due to its stability to the bandwidth hyperparameter in comparison to the Gaussian kernel, as well as its expressivity being equivalent to that of the neural tangent kernel of deep fully connected networks. However, unlike the Gaussian kernel, the Laplacian kernel is not separable. This poses challenges for techniques to approximate it, especially via the random Fourier features (RFF) methodology and its variants. In this work, we provide random features for the Laplacian kernel and its two generalizations: \matern kernel and the Exponential power kernel. We provide efficiently implementable schemes to sample weight matrices so that random features approximate these kernels. These weight matrices have a weakly coupled heavy-tailed randomness. Via numerical experiments on real datasets we demonstrate the efficacy of these random feature maps.
\end{abstract}


\section{Introduction}

Kernel machines are a classical family of models in machine learning. Certain architectures of neural networks \cite{lee2017deep,matthews2018gaussian,jacot2018neural} behave like kernel machines under a large width limit.

Recent work on kernel machines enabled with feature learning \cite{radhakrishnan2024mechanism}, which builds on the Laplacian kernel, has inspired many interpretable substitutes for deep neural networks on unstructured data. For example \cite{radhakrishnan2024synthetic} and \cite{aristoff2024fast} have demonstrated methodological advances in basic sciences.

From a theoretical standpoint, kernel machines provide a testbed for understanding many behaviours exhibited by deep neural networks \cite{belkin2018understand}. Recently, many complex neural phenomena been shown to be present in kernel models. Examples include emergence and grokking \cite{mallinar2024emergence}, deep neural collapse \cite{beaglehole2024average}, overparameterization \cite{simon2024more,ghosh2023universal}, benign overfitting \cite{mallinar2022benign}, and interpolation \cite{beaglehole2023inconsistency}, monotone improvement with model size \cite{abedsoltan2023toward}, and simplicity bias \cite{radhakrishnan2024mechanism} among others. Results on precise asymptotics \cite{mei2022generalization} have guided intuitions about neural networks \cite{ghorbani2021linearized,bartlett2021deep}.

The Laplacian kernel\footnote[2]{Let us clarify the nomenclature. \cite{rahimi2007random} refer to $e^{-\norm{x-z}_1}$ as the Laplacian kernel. However, following several recent works, we refer to \cref{def:l2laplacian} as the Laplacian kernel. We will refer to $e^{-\norm{x-z}_1}$ as the $\ell_1$-Laplacian kernel.
Note that unlike the $\ell_1$-Laplacian kernel, the Laplacian kernel \eqref{def:l2laplacian} is non-separable.}
\begin{align}
    K^{\textsf{L}}(x,z)=e^{-\norm{x-z}_M}\label{def:l2laplacian}
\end{align} is of particular interest, where $M$ is a positive definite matrix parameter. Even for $M=\gamma I_d$, the Reproducing Kernel Hilbert Space (RKHS) corresponding to this kernel is equivalent to the RKHSs corresponding to the Neural Tangent Kernel (NTK) for fully connected networks of any depth \cite{geifman2020similarity,chen2020deep}. Interpolation with \eqref{def:l2laplacian} can also be consistent in high dimensions \cite{rakhlin2019consistency,mallinar2022benign}.
Furthermore, unlike the Gaussian kernel, \eqref{def:l2laplacian} is excellent in practice for rapid exploratory data analysis since it is not sensitive to the bandwidth hyperparameter.

We consider the general anisotropic case with $M\neq \gamma I_d$ because it breaks the rotational symmetry. Kernel models trained via ridge regression can become degenerate for finite number of samples if the kernel function is rotationally symmetric \cite{karoui2013asymptotic}. Indeed, as shown in \cite{radhakrishnan2024mechanism}, the paramterization \eqref{def:l2laplacian} can be significantly more powerful if $M\neq \gamma I_d$ is chosen adaptively.

\subsection{Main contribution}

\begin{table}
    \centering
    \begin{tabular}{lcc}
    \toprule
    Kernel &  RFF ($W_{ij}$) & ORF ($S_{ii}$)\\
    & $x\mapsto\psi_p(Wx)$
    & $x\mapsto\psi_p(SQ\sqrt{M}x)$\\
    \midrule
    Gaussian 
    & $U_{ij}$ 
    & $q_i$
    \\
    $\ell_1$-Laplacian
    & $V_{ij}$
    & 
    \\
    \midrule
    Laplacian 
    & $U_{ij}/v_j$
    & $\sqrt{\ell_i}$
    \\
    Exp-power$(\alpha)$
    & $U_{ij}\cdot \sqrt{s_j}$
    & $q_i\cdot\sqrt{\omega_i}$
    \\
    Mattern$(\nu)$
    & $\sqrt{2\nu}\cdot U_{ij}/\tau_j$
    & $\sqrt{2\nu r_i}$
    \\
    \bottomrule
    \end{tabular}
    \caption{\label{tab:summary}\normalsize\textbf{Summary of random feature maps.}
    Schemes for sampling weight matrices to obtain random features. 
    $W\in\Real^{p\times d}$,  $S\in\Real^{p\times p}$ is a diagonal matrix. $\psi_p$ defined in \cref{eq:nonlinearity}.\\
    We require the weights to be sampled i.i.d. from their respective distributions $U_{i,:}\sim\mc N(0_d,M)$, $v_j\sim\mc N(0,1)$, $\omega_i,s_j\sim S\round{\tfrac{\alpha}{2}, 1,2\cos^{2/\alpha}(\tfrac{\pi\alpha}{4})},$ the univariate stable distribution. Further, $\tau_j \sim \chi(2\nu), q_i \sim \chi(d)$, use the Chi-distribution, and $ \ell_i \sim\textsf{BetaPrime}(\tfrac{d}{2}, \tfrac{1}{2}),$ and $r_i \sim\textsf{BetaPrime}(\tfrac{d}{2},\nu)$. See definitions of these distributions in \Cref{sec:prelim}.\\ 
    $Q\in\Real^{p\times d}$ is a uniformly distributed unitary matrix. $\sqrt{M}$ is the symmetric square root matrix of $M.$\\
    The schemes for Gaussian and $\ell_1$-Laplacian kernel are well-known. $V_{ij}\sim\text{Cauchy}(0,1)$.
    }
\end{table}

In this paper\footnote[3]{Code available at \href{https://github.com/parthe/torchkernels}{github.com/parthe/torchkernels}
} we provide explicit schemes to obtain random features that approximate the Laplacian kernel and its two generalizations: (i) the \matern kernel \cite{williams2006gaussian}, defined in \cref{def:matern} for $\nu>0$, and (ii) the Exponential power  kernel defined for $\alpha\in(0,2]$ as 
\begin{align}
    \label{def:exp-power}
    K^{\textsf{E}}_\alpha(x,z)=\exp\round{-\norm{x-z}_M^\alpha}
\end{align}
shown to be useful for practical tasks such as speech processing in 
\cite{Hui2018KernelMB}. We provide 2 schemes, RFF and ORF, for approximating all the 3 kernels. See \Cref{tab:summary} for a summary of how to generate these random features.

Via numerical experiments on 11 datasets we demonstrate the efficacy of the random features for kernel regression. 

We also show that kernel logistic regression, which yields far better calibrated models for classification than kernel ridge regression, is much easier to train with our random features than using the exact kernel function. 

Even for the isotropic case, $M=I_d$, the random features in \Cref{tab:summary} are coupled, i.e., not elementwise independent. This occurs because the kernel is not separable as a product of kernel functions over coordinates of the inputs. Yet, we are able to sample from these distributions efficiently.

\begin{figure}
\includegraphics[width=\columnwidth]{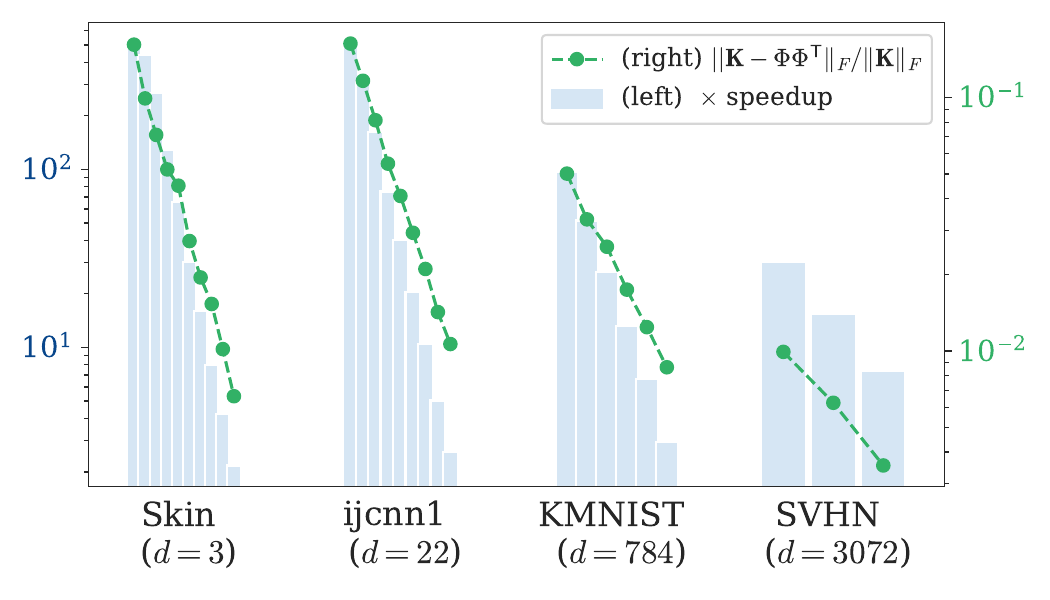}
\caption{\label{fig:speedup_by_datasets} Trade-off between computational cost and approximation error in evaluating Matern $\nu=4$ kernel across datasets with increasing dimensions and increasing $p$ for $n=10,000$ samples. $\Phi\in\Real^{n\times p}$ is the matrix of random features.
\vspace{-2ex}
}
\end{figure}

Recent work \cite{muyskens2024correspondence} has also highlighted the relation between the ReLU Neural Network Gaussian Process and the \matern-$\frac32$ kernel. However, evaluating the \matern kernel, can be expensive. Our schemes provide significant speed-ups (see \Cref{fig:speedup_by_datasets}). 

Formulae for Fourier transforms of the kernel functions, which are needed to understand weight distributions that yield appropriate random features, were known in niche references in applied mathematics. See \Cref{tab:fourier} for a summary. However, the application of these results to efficiently compute features has not been articulated explicitly in machine learning literature. In this paper we provide explicit feature maps (see \Cref{tab:summary}) along with experimental verification of their efficacy on real datasets.

\begin{remark}[Generalizations of Laplacian  kernels] 
By definition, \eqref{def:exp-power} becomes \eqref{def:l2laplacian} for $\alpha=1$. For $\nu=\frac12$, \eqref{def:matern} becomes \eqref{def:l2laplacian} as shown in \cite[eqns. 10.1.19 and 10.2.15]{abramowitz1948handbook}.
Note that $\tau_j\sim\xi(1)$ is a folded standard normal which along with symmetry of $U_{ij}$ acts like the normal distributed $v_j$. For $\nu=\frac12$, the variables $\ell_i$ and $r_i$ are identically distributed.

\end{remark}

In this paper we assume that for a fixed $p$, the nonlinear function $\psi_p:\Real\to\Real^2$ acts as
\begin{align}
\label{eq:nonlinearity}    \psi_p(t)=\tfrac{1}{\sqrt p}\round{\cos t,\sin t}\tran
\end{align}
and with some abuse of notation, for $u\in\Real^p$, we write $\psi_p(u)\in\Real^{2p}$ by stacking $\psi_p(u_i)$ vertically.

\section{Preliminaries}
\label{sec:prelim}

We consider the standard setting in supervised learning where we are provided with $n$ labeled training data points $\set{(x_i,y_i)\in\Real^d\times \Real}_{i=1}^n$ to learn a map $\wh f:\Real^d\mapsto\Real$.

A kernel $K:\Real^d\times \Real^d\to\Real$ is a bivariate, symmetric, positive definite function. Kernel machines are models
\begin{align}\label{eq:exact_kernel_model}
    \wh f(x)=\sum_{i=1}^n\wh\alpha_i K(x,x_i)
\end{align}
where $\wh\alpha_i$ are parameter trained on the labeled data.
Such functions are optimal for a wide class of non-parametric estimation problems of the form
\begin{align}\label{eq:representer_theorem}
\underset{f}{\text{minimize}}~ L(\set{(x_i,f(x_i),y_i)}_{i=1}^n) + R(\norm{f}_{\Hilbert_K})
\end{align}
for any loss function $L$ and any monotonically increasing function $R$, thanks to the representer theorem \cite{kimeldorf1971some,scholkopf2001generalized}. Here $\Hilbert_K$ is the RKHS corresponding to the kernel $K$. We refer the reader to \cite{steinwart2008support} for a detailed treatment on RKHS. 

If $K(x,z)=\inner{\phi(x),\phi(z)}_{\Hilbert'}$ for some feature map $ \phi:\Real^d\to\Hilbert'$, then we can write \cref{eq:exact_kernel_model} as $\wh f(x)=\inner{\phi(x),\wh\theta}_{\Hilbert'}$
where $\wh\theta\in\Hilbert'$ solves
\begin{align}\label{eq:featurized_glm}
    \underset{\theta\in\Hilbert'}{\text{min}}~ L(\set{(x_i,\inner{\phi(x_i),\theta}_{\Hilbert'},y_i)}_{i=1}^n) + R(\norm{\theta}_{\Hilbert'})
\end{align}
Furthermore, one can argue that $\wh\theta = \sum_{i=1}^n \wh\alpha_i\phi(x_i)$.

In this paper we focus on two types of feature maps $\phi(x)=\psi_p(Wx)$ and $\phi(x)=\psi_p(SQx)$ for  $W\in\Real^{p\times d}$, an orthogonal $Q\in\Real^{p\times d}$ and diagonal $S.$

We denote by $\Real^{d\times d}_\succ$ the space of $d\times d$ symmetric positive definite matrices.
\begin{definition}[Mahalanobis norm]
    For a symmetric positive definite matrix $M\in\Real^{d\times d}_{\succ}$, we denote the norm $\norm{u}_M:=\sqrt{u\tran Mu}$ for all $u\in\Real^d.$
\end{definition}
For a symmetric positive definite matrix $M$,  $\sqrt{\!M}$ denotes its unique symmetric square root.

We state a few distributions over $\Real_{\geq}$.

\begin{definition}[$\chi(k)$ and $\chi^2(k)$ distributions]\label{def:chi} The Chi distribution with $k>0$ degrees of freedom has density function
\begin{align}
    \frac{x^{k-1}e^{-x^2/2}}{\Gamma(\frac{k}2)\sqrt{2^{k}}} \quad \forall x\geq 0
\end{align}
where $\Gamma$ is the so-called Gamma function.
If $X\sim\chi(k)$ then $X^2\sim\chi^2(k).$
\end{definition}
These are exponential family distributions which can be sampled from efficiently. Furthermore, for $k\in\Natural$, we have $\norm{{w}}\sim\chi(k)$ if ${w}\sim\mc N(0,I_{k})$. 

\begin{definition}[\textsf{BetaPrime}$(\alpha,\beta)$ distribution]\label{def:beta-prime}
    The \textsf{BetaPrime} distribution with shape parameters $\alpha>0,$ and $ \beta>0$, has density function
    \begin{align*}
        f(x;\alpha,\beta) = \frac{x^{\alpha-1}(1+x)^{-\alpha-\beta}}{B(\alpha, \beta)}\quad \forall x\geq 0,
    \end{align*}
    where $B$ is the so-called Beta function.
\end{definition}

In Python we can  sample from this distribution using the \texttt{scipy.stats.betaprime} class. The \textsf{BetaPrime} distribution is the odds distribution of the Beta distribution, i.e., if $Z\sim \textsf{Beta}(\alpha, \beta)$, then $Y=\frac{Z}{1-Z}\sim \textsf{BetaPrime}(\alpha, \beta)$. 

\begin{definition}[Generalized Beta Prime ({\sf GBP}) distribution]\label{sample_GBP}
    If $u\overset{\sf i.i.d}{\sim} \textsf{BetaPrime}(\alpha,\beta)$, then $qu^{\frac{1}{p}}\overset{\sf i.i.d}{\sim} \textsf{GBP}(\alpha, \beta, p, q)$. The density function of this distribution is
    \begin{align*}
        \frac{p\left(\frac{x}{q}\right)^{\alpha p-1}}{qB(\alpha, \beta)\left(1+\left(\frac{x}{q}\right)^p\right)^{\alpha + \beta}}\quad\forall\,x\geq0.
    \end{align*}
\end{definition}

\begin{definition}[Multivariate $t_{2\nu}$-distribution]
    The centered multivariate $t$-distribution over $\Real^d$, with $2\nu$ degrees of freedom, and shape matrix $M \in \Real^{d\times d}_\succ$ has density function 
\begin{align*}
    \frac {\Gamma(\nu +\frac{d}2)}{\Gamma (\nu)\sqrt{\det(2\nu\pi M)}}\round{1+{\frac {\norm{x}^2_{M\inv}}{2\nu }}}^{-\nu-\frac{d}2}, \forall\, x\in\Real^d.
\end{align*} 
\end{definition}

The special case of the $2\nu=1$ yields the multivariate Cauchy distribution.

\subsection{Elliptically contoured $\alpha$-stable distribution}

The elliptically contoured $\alpha$-stable distribution is a special class of multivariate $\alpha$-stable distributions. To introduce the multivariate stable distribution, we first need to define the univariate stable distribution. We direct the reader to \cite{nolan2020univariate} for a detailed treatment on the stable distribution, and \cite{samoradnitsky2017stable} for the multivariate setting.

A stable distribution is one that is closed under positive linear combinations, upto a change of location and scale, i.e., if we take two independent realizations $X_1, X_2$ of a random variable $X$ and positive constants $a>0, b>0$, the random variable $aX_1 + bX_2$ has the same distribution as $cX+d$ for some constants $c>0, d\in \Real$.

It is often convenient to use a more concrete definition of the stable distribution such as the one given below. A proof of the equivalence of these definitions is given in \cite[Ch. 3]{nolan2020univariate}

The general discussion on stable distributions has a location parameter $\mu$. In this work we assume the location is $0$ and hence omit this parameter in our discussion. We consider the 3 parameter family $S(\alpha,\beta,\sigma)$ defined below. The stable distribution, unlike most distributions, is defined via its characteristic function. The density function cannot be expressed analytically for general $\alpha, \beta, \sigma$.
\begin{definition}[Univariate stable distribution \cite{nolan2020univariate}]\label{def:univar_stable_dist}
    The stable distribution $S(\alpha,\beta,\sigma)$ with shape parameters $\alpha\in(0,2]$, and $\beta\in[-1,1]$ and scale parameter $\sigma>0$ has the characteristic function 
    \begin{align}
        &\exp \left(-|\sigma t|^{\alpha }\left(1-j\beta \operatorname {sgn}(t)\Phi(t) \right)\right) \quad\forall\,t\in\Real\label{def:stable}\\
        &\Phi(t) =\begin{cases}\tan \left({\frac {\pi \alpha }{2}}\right)&\alpha \neq 1\\-{\frac {2}{\pi }}\log |t|&\alpha =1\end{cases}.\nonumber
    \end{align}
\end{definition}
The above definition is referred to as the `Nolan's 0' parametrization \cite{nolan2020univariate}.

 The multivariate $\alpha$-stable distribution is defined in terms of it's projections. 
\begin{definition}[Multivariate $\alpha$-stable distribution] A $\Real^d$-valued random variable $X$ is said to have a multivariate $\alpha$-stable distribution if, for any $u \in \Real^d$, $u\tran X\sim S(\alpha,\beta_u,\sigma_u)$ for some $\beta_u\in[-1,1],$ and $\sigma_u>0$. Note that $\alpha$ does not depend on $u.$
\end{definition}
\begin{definition}(Elliptically contoured $\alpha$-Stable distribution) \cite[Sec.3]{nolan2005multivariate}\label{def:eca-stable}
    An $\Real^d$-valued random variable $X$ is said to be elliptically contoured $\alpha$-stable if $u\tran X\sim S(\alpha,0,\norm{u}_M)$, for some positive definite shape matrix $M\in\Real^{d\times d}_{\succ}$. Its characteristic function is given by
    \begin{align*}
        \Exp \exp\round{j\inner{u,X}} = \exp\round{-\norm{u}_M^\alpha}.
    \end{align*}
\end{definition}

\subsection{The \texorpdfstring{\matern}{matern} kernel}

The Laplacian and the Gaussian kernel are members of several families of kernels. One very natural extension comes from the observation that the Cauchy distribution is the $t$-distribution with $1$ degree of freedom ($2\nu = 1$). Further, as the number of degrees of freedom rise to infinity ($2\nu \rightarrow \infty$), $t$-distribution approaches the Gaussian. Using these extremes, Fourier transforms of all intermediate $t$-distributions also define shift-invariant kernels, together studied as the \matern kernel with parameter $\nu$. A Gaussian process with \matern covariance is $\lceil \nu \rceil-1$ times differentiable in the mean-squared sense \cite[Sec. 4.2.1]{williams2006gaussian}.

\begin{definition}[\texorpdfstring{\matern}{matern} kernel]
For $\nu>0$, 
\begin{align}
    &K^{\textsf{M}}_\nu(x,z) := \kappa_\nu(x-z)\nonumber\\
    &\kappa_\nu(\Delta):=\tfrac {2^{1-\nu }}{\Gamma (\nu )} \round{\sqrt{2\nu }\norm{\Delta}_M}^\nu J_{\nu }\round{\sqrt {2\nu }\norm{\Delta}_M}\label{def:matern}
\end{align} where $M\in \Real^{d\times d}_\succ$ is a scale matrix and $\nu$ is the shape factor and $J_\nu$ is the modified Bessel function of the second kind. 
\end{definition}
For certain values of $\nu$, the \matern kernel can be simplified:
\begin{align*}
    \kappa_{1/2}(\Delta)\! &=\! e^{-\norm{\Delta}_M},\qquad
    \kappa_{\infty}(\Delta) = e^{-\frac{1}{2}\norm{\Delta}_M^2},\\
    \kappa_{3/2}(\Delta)\! &=\! \left(1+\sqrt{3}\norm{\Delta}_M\right)e^{-\sqrt{3}\norm{\Delta}_M}, \text{ and}\\
    \kappa_{5/2}(\Delta)\! &=\! \left(1\!+\!\sqrt{5}\norm{\Delta}_M\!+\!\tfrac53\norm{\Delta}_M^2\right)e^{-\sqrt{5}\norm{\Delta}_M}.
\end{align*}
In general, for $ \nu \in \curly{n+\frac12}_{n \in\mathbb{N}}$, one can simplify the \matern kernel in terms of a polynomial times an exponential, similar to the examples above. Another special case is $n=\infty$ where the \matern is exactly equal to the Gaussian. The \texttt{sklearn} package \cite{scikit-learn} states that for $\nu\notin\set{\frac12, \frac32, \frac52,\infty}$ one incurs a considerably high computational cost ($\sim10\times$) since it requires the evaluation of the modified Bessel function. Hence, random features are more effective for other values of $\nu$. See \Cref{fig:speedup_by_datasets} for a tradeoff between approximation error and computational speedup for datasets of different dimensions.

\section{Background on random features}
The key idea of RFF, proposed in \cite{rahimi2007random}, comes from a classical theorem in harmonic analysis due to Bochner \cite[Thm.1.4.3]{rudin2017fourier}. Bochner's theorem states that any function $\kappa$ is positive definite if and only if it is the characteristic function of a random variable (upto a normalizing constant), i.e., the Fourier transform of $\kappa$ is a measure. This enables writing any shift invariant kernel as \begin{align}
K(x,z)=\kappa(x-z)&=c_\kappa\cdot\Exp_w e^{j\inner{w,x-z}},
\end{align}
for a normalizing constant $c_\kappa$. Here $w$ is a random variable with measure proportional to the Fourier transform of $\kappa$, normalized to have $\Prob(w\in\Real^d)=1.$

\subsection{Random Fourier features (RFF)}
We present below the so-called SinCos RFF maps from \cite{sutherland2015error} due to their lower approximation error over the RFF scheme proposed in \cite{rahimi2007random}.
\begin{proposition}\label{prop:rff_proof}
    Let $\set{w_i}_{i=1}^p$ be i.i.d. samples from a distribution over $\Real^d$ whose characteristic function is $\frac1{c_\kappa}\kappa$. Let $\psi_p$ be an elementwise nonlinearity defined in \cref{eq:nonlinearity}, and suppose the rows of $W\in\Real^{p\times d}$ are $w_i$. Then
    \begin{align}
        \lim_{p\to\infty}
        \inner{\psi_p(Wx),\psi_p(Wz)}=\kappa(x-z)=K(x,z)
    \end{align}
\end{proposition}
To prove the above result, observe that $\inner{\psi_p(Wx),\psi_p(Wz)}_{\Real^{2p}}$ equals
\begin{align*}
&\frac{1}p\sum_{i=1}^p\cos(w_i\tran x)\cos(w_i\tran z)+\sin(w_i\tran x)\sin(w_i\tran z)\xrightarrow{p\to\infty}\\
&c_\kappa\Exp_w \cos(w(x-z))= c_\kappa \Exp_w e^{j\inner{w,x-z}}=\kappa(x-z)
\end{align*}
where the limit holds almost surely, by strong law. For a non-asymptotic characterization for finite $p$, please refer to \cite[Prop. 1]{sutherland2015error}.
\subsection{Orthogonal random features (ORF)}
Proposed in \cite{yu2016orthogonal}, ORF provides a structured method for creating $ W$, leveraging the distribution of $\norm{w}$. 
\begin{proposition}\label{prop:orf_proof}
    Let $p$ be an integer multiple of $d$, and $Q\in\Real^{p\times d}$ be obtained by stacking $\frac{p}{d}$ independent samples from uniform distribution over $d\times d$ unitary matrices.
    Let $w$ be a $\Real^d$ valued random variable, whose characteristic function is $\frac1{c_\kappa}\kappa$, for some rotationally invariant function $\kappa$, i.e., $\kappa(V\Delta)=\kappa(\Delta)$ for all unitary matrices $V\in\Real^{d\times d}$. Let $S\in\Real^{p\times p}$ be a diagonal matrix with entries
    $\set{S_{ii}}_{i=1}^p$ sampled i.i.d. from the distribution of $\norm{w}$. Let $\psi_p$ be an elementwise nonlinearity defined in \cref{eq:nonlinearity}. Then
    \begin{align}
        \lim_{p\to\infty}\inner{\psi_p(SQx),\psi_p(SQz)}=\kappa(x-z).
    \end{align}
\end{proposition}
\cite{yu2016orthogonal} provide a result for the Gaussian kernel. Their argument can be extended to general rotationally invariant kernels as follows. If $\kappa$ is rotationally invariant, then one can show that the distribution of $w$ is rotationally symmetric, whereby $\norm{w}$ and $\frac{w}{\norm{w}}$ are independent random variables. Furthermore, $\frac{w}{\norm{w}}$ is uniformly distributed over $\mathbb{S}^{d-1}$, the unit sphere in $d$ dimensions. Thus, $SQ$ is a reparameterization of writing $w=\norm{w}\cdot\frac{w}{\norm{w}}$; the rows of $Q$ are distributed like $\frac{w}{\norm{w}}$, whereas the diagonal elements of $S$ are distributed like $\norm{w}$. Together, $SQ$ is distributed like $W$ from \Cref{prop:rff_proof}.
 \paragraph{Extending ORF to the anisotropic case.} The above result describes how to determine random features for the kernel with $M=I_d$. The generalization to $M \in \Real^{d\times d}_\succ,  M\neq I$ can be obtained by using $\norm{u}_M = \norm{\sqrt{M}u}$,
 where $\sqrt{M}$ is the unique symmetric positive definite square root matrix of $M.$ Thus $K_M(x,z)=K_{I_d}(\sqrt{M}x,\sqrt{M}z)$. 

We provide a brief overview of works related to random features.

\subsection{Prior work on random features}

After the introduction of RFF in \cite{rahimi2007random}, several improvements have been suggested. We focus primarily on the algorithmic improvements. See \cite{liu2021random} for a comprehensive survey. 

For the Gaussian kernel, a long line of work improved upon the vanilla RFF formulation to either reduce space and time complexity or improve approximation and generalization error. For example, Fastfood \cite{Fastfood} and it's generalization $\mathcal{P}$-model \cite{P_model} utilize Hadamard matrices to speed up computation of $x\mapsto Wx$. An alternate approach was suggested in \cite{scrf} which utilized signed circulant matrices to generate features. \cite{NRFF} argues that  normalizing the inputs leads to gains in approximation and generalization performance due to restricting the diameter of the data.

\begin{remark}[Other random feature maps]
    Techniques like ORF, \cite{yu2016orthogonal}, use an orthogonal rotation matrix along with a radial distribution multiplier. This is shown to be unbiased and has a lower variance than vanilla RFF. In \cite{yu2016orthogonal} they also introduce ORF-prime, a version of ORF, with a constant radial component, which works well in practice for the Gaussian distribution. However, recently in \cite[Thm. 2]{demni2024orthogonal}, this was shown to actually approximate the normalized Bessel function of the first kind, which is different from the Gaussian. Structured ORF (SORF) \cite{yu2016orthogonal} uses products of pairs of sign-flipping and Hadamard matrices ($HD$ blocks) to approximate $W$. SORF uses ORF-prime and replaces the orthogonal matrix with products of $HD$ blocks. However, whether SORF also demonstrates the  bias shown in \cite{demni2024orthogonal} is not known. For the above reasons we do not include SORF and ORF-prime in our discussion. \cite{Bojarski2016StructuredAA} extends this idea using random spinners. \cite{ROM_} generalizes SORF by using an arbitrary number of sign-flipping and Hadamard matrix blocks and also provides intution for why 3 blocks work well in practice. 
\end{remark}

Quadrature rules approximate the Fourier transform integral \cite{Gauss_quad}, \cite{Quad_based_feats}, however, these works assume separability of the integral which is available in case of the Gaussian and the $\ell_1$-Laplacian kernel. \cite{Bach_equivalence} showed the equivalence between Random Fourier Features and kernel quadrature rules. While quadrature based approaches are more general, they too assume separability, and subgaussianity. 

Random features for dot product kernels introduced 
in \cite{pmlr-v22-kar12}, and were generalized in \cite{wacker2024improved} to include sketching. Dot product kernels rely on the Mclaurin series expansion, which assumes existence of all derivatives, an assumption not satisfied by \eqref{def:l2laplacian}.

\section{Fourier transform of \texorpdfstring{$\kappa_{1/2}$ and $\kappa_{\nu}$}{kappa}}

\begin{table*}
\centering
    \begin{tabular}{lclcl}
    \toprule
    Kernel & $\kappa(\Delta) $ & 
    {Fourier transform} &
    $\mc F\set{\kappa}(w)$
    &
    Reference \\
    \midrule
    Gaussian 
    & $e^{-\frac{1}{2}\norm{\Delta}_M^2}$
    & Gaussian
    & $e^{-\frac1{2}\norm{w}_{M\inv}^2}$
    \\
    Laplacian 
    & $e^{-\norm{\Delta}_M}$
    & Cauchy
    & $\round{1+\norm{w}_{M\inv}^2}^{-\frac{d+1}{2}}$
    & \cite[Theorem 1.14]{Stein1971}
    \\
    Exp-power 
    & $e^{-\norm{\Delta}_M^\alpha}$
    & $\alpha$-stable$^\dagger$
    & \cref{def:stable}
    & \cite[Proposition 2.5.8]{samoradnitsky2017stable}
    \\
    \matern &
    \cref{def:matern}
    & $t_{2\nu}$-distribution
    & $\round{1+\frac{\norm{w}_{M\inv}^2}{2\nu}}^{-\frac{d+2\nu}{2}}$ 
    & \cite[Theorem 3.1]{joarder1996characteristic}\\
    \bottomrule
    \end{tabular}    \caption{\normalsize\label{tab:fourier}\textbf{Fourier transforms} (up to multiplicative constants) of some shift-invariant kernel functions $K(x,z)=\kappa(x-z)$ where $\kappa:\Real^d\to\Real$, and $M\in\Real^{d\times d}_{\succ}$ is a positive definite matrix.\\
    $^\dagger$The precise name is elliptically-contoured $\alpha-$stable distribution.}

\end{table*}

To identify the distribution of weights in the random features, we must find out the fourier transform of the $\kappa$ function.
\begin{align*}
    \mc F\set{\kappa}(w):=\int_{\Real^d} \kappa(\Delta)e^{-j\inner{w,\Delta}}\dif\Delta
\end{align*}
Note that the anisotropic case $M\neq \sigma^2 I_d$, can be dealt with easily. $\mc F\set{\kappa}$ depends on the dual norm of the norm in $\kappa$.

\begin{lemma}\cite[Thm. 1.14]{Stein1971} 
\label{lemma:fourier_laplace}
    Fourier transform the $\e^{-\norm{\Delta}_M}, M\in\Real^{d\times d}_\succ$ is the \textsf{Cauchy}$(0_d, M)$ distribution.
    \end{lemma}
The interested reader can see an elegant proof by \cite{Stein1971} in the \Cref{appendix:proofs}.

In case of the Laplacian kernel, clever manipulation enabled us to obtain the Fourier transform. This is difficult to do for the general \matern kernel primarily due to the presence of a Bessel Function. \cite{geom_random} also provide closed forms for the fourier transform of the \matern kernel, without mentioning the $t$-distribution.

\begin{proposition}\texorpdfstring{\cite[Thm. 3.1]{joarder1996characteristic}}{Thm 3.1 Joarder}\label{prop:fourier_t}
    The characteristic function of the $t_{2\nu}$-distribution is 
    \begin{align*}
        \frac{\norm{\sqrt{2\nu } w}_M^{\nu}}{2^{\nu-1}\Gamma(\nu)}J_{\nu}\left(\norm{\sqrt{2\nu } w}_M\right),\qquad\forall\,w\in\Real^d
    \end{align*}
where
$M\in\Real^{d\times d}_{\succ}$ and $ J_{\nu}:\Real\to\Real$ is the modified Bessel function of the second kind of order $\nu$.
\end{proposition}
For a proof of this theorem, we refer the reader to \cite{Kotz_Nadarajah_2004}, and \cite{characteristic_t} for a much simpler proof of the one-dimensional case.

\cite{sutradhar1986characteristic} also studied the characteristic function of the $t$-distribution, where they formulated it in terms of separate infinite series for odd, even and fractional degrees of freedom. \cite{joarder1996characteristic} showed the relationship in terms of the Bessel function. Note that this form is also known in \cite{geom_random}, though not operationalized to obtain random features.

\section{Sampling from heavy-tailed anisotropic distributions using elementary distributions}
We state a few centered\footnote{Note these distributions are symmetric about the origin. Even though the mean (first moment) may not exist for these distributions, $\Prob(X\in A)=\Prob(X\in -A)$ for all measurable sets $A\subset \Real^d$. Here $-A=\set{-x\mid x\in A}.$} multivariate distributions with non-identity shape matrix $M\in\Real^{d\times d}_{\succ}$. We discuss two ways to sample these distributions.
\begin{enumerate}
    \item Using correlated multivariate Gaussians: Sample $\mc N(0, M)$ and scale it appropriately.
    \item Using the norm distribution: Sample a uniformly random unit vector, apply $\sqrt{M}$, and scale it appropriately.
\end{enumerate}

\subsection{Sampling from elliptically contoured \texorpdfstring{$\alpha$}{alpha}-stable distribution}

Recall the discussion on elliptically contoured $\alpha$-stable distribution.
\begin{proposition}\cite[Prop. 2.5.2]{samoradnitsky2017stable}{} 
    Let $M\in\Real^{d\times d}_\succ$, and consider two independent random variables $G \sim \mc N(0,M)$, and $A \sim S(\frac{\alpha}{2}, 1, 1)$. Then $\sqrt{\!A}G$ is an elliptically contoured $\alpha$-stable distribution with shape matrix $M$ and has characteristic function 
    \begin{align*}
        \Exp e^{j\inner{u,\sqrt{\!A}G}} = \exp\round{-\round{\tfrac{1}{2}}^{\alpha/2} \sec\frac{\pi\alpha}{4}\norm{u}_M^\alpha}
    \end{align*}
\end{proposition}
The following corollary, which chooses a particular value of $\sigma$ to cancel the exponent, provides a direct method of generating i.i.d. samples from the required stable distribution.
\begin{corollary}\label{coro:sample_stable}
    If $\alpha\in(0,2)$, $A \sim S\round{\frac{\alpha}{2}, 1, 2\cos^{\frac2\alpha}\frac{\pi\alpha}4}$ and $G\sim \mc N(0,M)$, $ M\!\in\Real^{d\times d}_{\succ}$, then 
    \begin{align*}
        \Exp \exp\round{j\inner{u,\sqrt{\!A}G}}=\exp\round{-\norm{u}_M^\alpha}
    \end{align*}
\end{corollary}
Note that, the above results have allowed us to reduce the problem of sampling from a multivariate stable distribution to that of a univariate $\alpha$-stable distribution. 

Many methods are available, to sample from univariate $\alpha$-stable distribution. 
The CMS algorithm \cite{chambers1976method}, provided as \Cref{algo:cms} in \Cref{appendix:expts}, enables sampling from $S(\alpha, \beta, \sigma)$, the proof of which was provided in \cite{weron1996chambers}. 

\subsection{Sampling from multivariate Students \texorpdfstring{$t$}{t}-distribution}
The following lemma is the constructive definition of the multivariate $t$-distribution and simultaneously serves as a sampling algorithm.
\begin{lemma}\label{lemma:sample_t}
     Let $M\in\Real^{d\times d}_\succ$ and $\nu>0$. If ${u} \sim \mc N(0,M)$ and $v \sim \chi^2_{2\nu}$ and $u, v$ are independent then, $u \sqrt{\frac{{2\nu}}{v}}$ has the multivariate $t$-distribution with ${2\nu}$ degrees of freedom and shape matrix $M\in\Real^{d\times d}_{\succ}$.
\end{lemma}

The following lemma allows us to sample from the radial measure of the isotropic multivariate $t$-distribution.
\begin{lemma}\label{lemma:sample_norm_of_t}
If ${w}$ is a multivariate $t$-distributed random variable with $2\nu$ degrees of freedom and shape matrix $\sigma^2 I_d$, then we have $\norm{{w}}\sim\textsf{GBP}(\frac{d}{2}, \nu, 2, \sigma\sqrt{2\nu})$.
\end{lemma}
We provide the proof in \Cref{appendix:proofs}.

\subsection{Sampling from multivariate Cauchy distribution}
Notice that the Cauchy distribution is a special case of the $t$-distribution, thus a result similar to \cref{lemma:sample_t} is applicable in case of the Cauchy distribution. 
\begin{lemma}\label{lemma:sample_cauchy}
    Let $M\in\Real^{d\times d}_\succ$ and consider independent random variables $u \sim \mc N(0,M)$ and $v \sim \mc N(0,1)$. Then, $u/v \sim \textsf{Cauchy}(0_d, M)$.
\end{lemma}
Since $\textsf{Cauchy}(0_d, M)$ is symmetric about the origin, dividing by a $\chi(1)$ random variable, which has the folded standard normal distribution, is equivalent to dividing by a standard normal random variable.

The following claim follows immediately from \Cref{lemma:sample_norm_of_t}, since Cauchy distribution is the $t$-distribution with $\nu=\frac12$.
\begin{corollary}\label{lemma:sample_norm_of_cauchy}
    If ${w}\sim\textsf{Cauchy}(0_d,\sigma^2 I_d)$, then we have $\norm{{w}}\sim \textsf{GBP}(\frac{d}{2}, \frac{1}{2}, 2, \sigma)$.
\end{corollary}

\section{Main results}
Recall definitions of $S(\alpha,\beta,\sigma)$, and $\chi(2\nu)$ from \Cref{sec:prelim}.
\begin{theorem}[RFF for the Laplacian, Exponential-power and \texorpdfstring{\matern}{matern} kernels]\label{thm:RFF} Suppose $\nu>0$, $\alpha\in(0,2]$ and $M\in\Real^{d\times d}_{\succ}$. Consider sets of independent random variables $\set{u_i}_{i=1}^p\overset{\sf i.i.d.}\sim \mc N(0,M)$, $\set{v_i}_{i=1}^p\overset{\sf i.i.d.}\sim \mc N(0,1),$ $\set{s_i}_{i=1}^p\overset{\sf i.i.d.}\sim S\round{\frac\alpha2,1,2\cos^{2/\alpha}\frac{\pi\alpha}{4}}$, and $\set{\tau_i}_{i=1}^p\overset{\sf i.i.d.}\sim \chi(2\nu)$, and construct matrices $W^{\sf L}=\round{\frac{u_i}{v_i}}_{i}$, $W^{\sf E}=\round{u_i\cdot\sqrt{s_i}}_{i}$, and $W^{\sf M}=\round{\frac{u_i}{\tau_i}}_{i}$ by stacking the $\Real^d$ vectors along the rows. Then for all $x,z\in\Real^d$, for all $\eta\in\set{\sf L, E, M}$, we have almost surely,
\begin{align*}
&\lim_{p\to\infty}\inner{\psi_p(W^\eta x),\psi_p(W^\eta z)}_{\Real^{2p}}=K^\eta(x,z).
\end{align*}
\end{theorem}
\begin{proof}[Proof sketch.]
    Note that if $\bm{u_i}\sim \mc N(0, M)$ and $v_i\sim\mc N(0,1)$ then $\frac{u_i}{v_i}\sim\textsf{Cauchy}(0_d,M)$ due to \Cref{lemma:sample_cauchy}. Note that $\textsf{Cauchy}(0_d,M)$ is the Fourier transform for $e^{-\norm{\Delta}_M}$ thanks to \Cref{lemma:fourier_laplace}. The claim for $K^{\sf L}$ follows via \Cref{prop:rff_proof}. 
    Using a similar argument equipped with \Cref{lemma:sample_t} and \Cref{prop:fourier_t}, the claim for $K^{\sf M}$ follows.

    The construction of $W^{\sf E}$ is such that each row is a sample from a multivariate $\alpha$-Stable distribution, due to \Cref{coro:sample_stable}. Which, by definition has the desired characteristic function. The claim follows immediately by \Cref{prop:rff_proof}.
\end{proof}

Recall the definition of the generalized beta prime distribution, i.e., \textsf{GBP}$(\alpha, \beta, p, q)$.

\begin{theorem}[ORF for the Laplacian, Exponential-power and \texorpdfstring{\matern}{matern} kernels]\label{thm:ORF}
   Let $\frac{p}{d}\in\Natural, M\in\Real^{d\times d}_{\succ}$, and let $Q$ be a unitary matrix of size $p\times d$ obtained by vertically stacking $\frac{p}{d}$ independent uniformly distributed $d\times d$ matrices. Let $S^{\sf L},S^{\sf E},S^{\sf M} \in \Real^{p\times p}$ be diagonal matrices, with entries sampled independently as $\set{S^{\sf L}_{ii}}_{i=1}^p\overset{\sf i.i.d.}\sim\textsf{GBP}\left(\frac{d}{2},\frac{1}{2},2,1\right)$, $\set{S^{\sf E}_{ii}}_{i=1}^p\overset{\sf i.i.d.}\sim S\round{\frac{\alpha}{2}, 1, 2\cos^{\frac2\alpha}\frac{\pi\alpha}4}$, and $\set{S^{\sf M}_{ii}}_{i=1}^p\overset{\sf i.i.d.}\sim\textsf{GBP}\left(\frac{d}{2},\nu,2,\sqrt{2\nu}\right)$. 
   Then, for any $x,z\in\Real^d$, for all $\eta\in\set{\sf L, E, M}$, we have almost surely
   \begin{align*}
&\lim_{p\to\infty}\inner{\psi_p(S^\eta Q\sqrt{M} x),\psi_p(S^\eta Q\sqrt{M} z)}_{\Real^{2p}}=K^\eta (x,z).
\end{align*}
\end{theorem}
The proof follows from \Cref{prop:orf_proof} via arguments similar to \Cref{thm:RFF}. However, we now need to work with distributions of $\norm{w}$ where the characteristic function of $w$ is $\frac1{c_\kappa}\kappa$. These distributions have been derived in \Cref{lemma:sample_norm_of_cauchy}, and \Cref{lemma:sample_norm_of_t}.

\section{Numerical experiments}
Our experiments cover $6=2\times3$ configurations: 2 types of random features (RFF/ORF) and 3 kernels (Laplacian/ \matern /Exponential-power). In the main text, we have shown only limited experiments. The appendix contains the remaining configurations. Our experiments are for the isotropic case of $M=I_d$.

\begin{figure*}[t]
\includegraphics[width=0.95\textwidth]{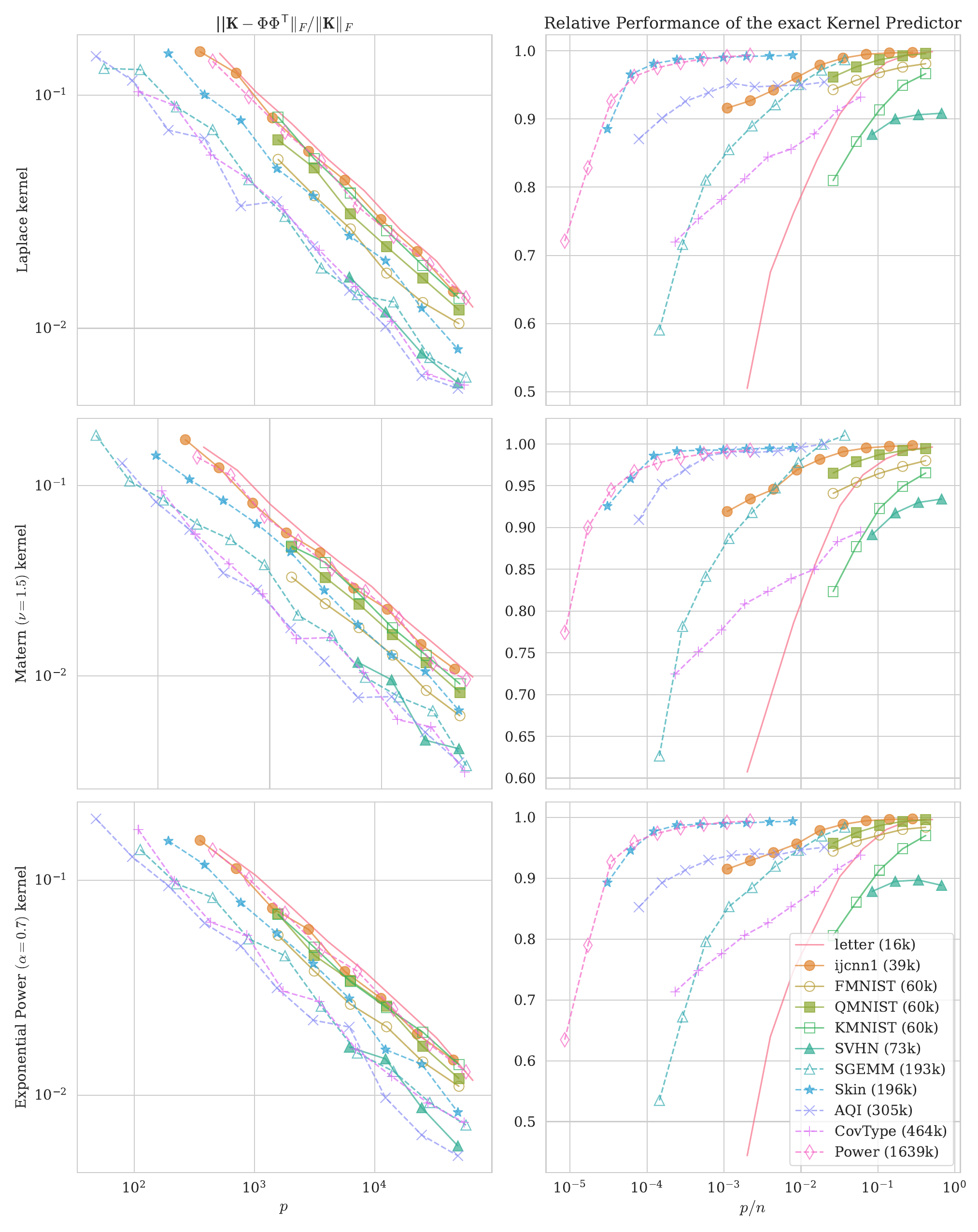}
\caption{\normalsize\label{fig:ORF_eval}Evaluation of ORF. \textbf{(Left)}: relative approximation error for the kernel matrix for various datasets in frobenius norm, for $n=10,000$ samples. Error measured in other norms (operator, nuclear) are mentioned in \Cref{appendix:expts} (See Figures \ref{fig:approx_op_norm}, \ref{fig:approx_nuc_norm}). Here, $\Phi\in\Real^{n\times p}$ is the matrix of random features. \textbf{(Right)}: performance of random features predictor in comparison to exact solution via EigenPro2 \cite{ma2019kernel}. The unnormalized numbers are available in \Cref{tab:krr}. See \Cref{fig:RFF_eval} for similar evaluation of RFF.}
\end{figure*}

\subsection{Logistic regression with random features}

We compare the performance of solving \cref{eq:representer_theorem} with $L$ as the empirical cross-entropy loss with a ridge penalty $R(t)=\half t^2$. We compare  the two paramterizations (i) the exact kernel parameterization in  \cref{eq:exact_kernel_model} which involves a search over $\Real^n,$ and (ii) the RFF parameterization $\phi(x)=\psi_p(Wx)$ in \cref{eq:featurized_glm}.

\begin{table}[t]
    \centering
    \caption{\label{tab:ece}Comparison of expected calibration error (ECE, lower is better) of kernel ridge regression and kernel logistic regression and its approximate versions. Number of random features $p=10,000$ and $n$ denotes the training data size. We use $\alpha=0.7$ for Exponential-power kernel. All models have nearly the same accuracy. But ORF with logistic regression has well callibrated models. The ORF with logisitic regression runs upto $50\times$ faster.}
    \begin{tabular}{llccc}
    \toprule
        Dataset & Kernel &
        LS loss &   \multicolumn{2}{c}{Logistic loss}\\
        \cmidrule(lr){4-5}
        &
        & exact
        & ORF (ours)
        & exact \\
        \midrule
        FMNIST & Laplacian & 0.70 & 0.021 & \textbf{0.011} \\
        $d=784$ & Exp-power & 0.68 & 0.026&\textbf{0.009}\\
        $n=$60\,{\rm k} & Mattern-$\frac32$   &0.69 & \textbf{0.006}& 0.013 \\
        \midrule
        KMNIST & Laplacian   & 0.74 & \textbf{0.022} & 0.027 \\
        $d=784$ & Exp-power  & 0.72& \textbf{0.018}& 0.023\\
        $n=$ 60\,{\rm k} & Mattern-$\frac32$  & 0.71& \textbf{0.024} & \textbf{0.024}\\
        \midrule
        QMNIST & Laplacian  & 0.77 & \textbf{0.013} & 0.018 \\
        $d=784$ & Exp-power & 0.76& \textbf{0.014}& \textbf{0.014}\\
        $n=$60\,{\rm k} & Mattern-$\frac32$  & 0.76& \textbf{0.013}& 0.017\\
        \midrule
        SVHN & Laplacian   & 0.64 & \textbf{0.024} & 0.145\\
        $d=3072$ & Exp-power & 0.63& \textbf{0.015} & 0.140 \\
        $n=$73\,{\rm k} & Mattern-$\frac32$ & 0.63& \textbf{0.111}& 0.140\\
    \bottomrule
    \vspace{-5ex}
    \end{tabular}
\end{table}

\section{Discussion and outlook}

In this paper we presented 2 schemes, RFF and ORF, to approximate the Laplacian kernel and its two generalizations -- \matern and Exponential-power kernels. The Laplacian kernel is radial and hence weakly non-separable. This leads to the random weight matrix for the random features to be coupled along its rows. 
We provide efficient ways to sample the random weights using simpler distributions. 

Our kernel evaluations are very fast in comparison to standard techniques in \texttt{sklearn}. Similar to the paradigm of neural networks, our approach also allows for scaling the training data arbitrarily from the model size. This opens the door for using best practices from deep learning such as data augmentation to kernel machines. 

We also demonstrate that kernel regression with the logistic loss yields better calibrated models. These models can be hard to train beyond the square-loss. In such scenarios, we are better off using random features since the approximation errors are quite low.

In practice, sampling from very heavy tailed distributions can be prone to errors. This is especially the case for the Exponential-power kernel with $0<\alpha<0.3$. Since these distributions are very flat (i.e. have very fat tails) sampling using floats leads to rounding off errors. This is because given a memory budget (e.g. 32 bits, \cite[IEEE 754]{IEEE_754}) one part is allocated towards storing the fraction (23 bits), another towards the exponent (8 bits) and one bit is typically reserved for the sign. Thus, sampling numbers farther away from the origin leads to larger rounding off errors. Thus one is forced to use a higher precision float when sampling from such distributions. Analyzing the effect of these rounding off errors on both kernel approximation as well as generalization error remains an object of further study. 

\subsection*{Acknowledgements}
PP acknowledges support from Schmidt Sciences and the DST INSPIRE Faculty Fellowship.

\bibliography{ref}

\begin{thebibliography}{59}
\providecommand{\natexlab}[1]{#1}
\providecommand{\url}[1]{\texttt{#1}}
\expandafter\ifx\csname urlstyle\endcsname\relax
  \providecommand{\doi}[1]{doi: #1}\else
  \providecommand{\doi}{doi: \begingroup \urlstyle{rm}\Url}\fi

\bibitem[IEE(2019)]{IEEE_754}
Ieee standard for floating-point arithmetic.
\newblock \emph{IEEE Std 754-2019 (Revision of IEEE 754-2008)}, pp.\  1--84, 2019.
\newblock \doi{10.1109/IEEESTD.2019.8766229}.

\bibitem[Abedsoltan et~al.(2023)Abedsoltan, Belkin, and Pandit]{abedsoltan2023toward}
Abedsoltan, A., Belkin, M., and Pandit, P.
\newblock Toward large kernel models.
\newblock In \emph{International Conference on Machine Learning}, pp.\  61--78. PMLR, 2023.

\bibitem[Abramowitz \& Stegun(1948)Abramowitz and Stegun]{abramowitz1948handbook}
Abramowitz, M. and Stegun, I.~A.
\newblock \emph{Handbook of mathematical functions with formulas, graphs, and mathematical tables}, volume~55.
\newblock US Government printing office, 1948.

\bibitem[Aristoff et~al.(2024)Aristoff, Johnson, Simpson, and Webber]{aristoff2024fast}
Aristoff, D., Johnson, M., Simpson, G., and Webber, R.~J.
\newblock The fast committor machine: Interpretable prediction with kernels.
\newblock \emph{The Journal of chemical physics}, 161\penalty0 (8), 2024.

\bibitem[Bach(2017)]{Bach_equivalence}
Bach, F.
\newblock On the equivalence between kernel quadrature rules and random feature expansions.
\newblock \emph{J. Mach. Learn. Res.}, 18\penalty0 (1):\penalty0 714–751, January 2017.
\newblock ISSN 1532-4435.

\bibitem[Bartlett et~al.(2021)Bartlett, Montanari, and Rakhlin]{bartlett2021deep}
Bartlett, P.~L., Montanari, A., and Rakhlin, A.
\newblock Deep learning: a statistical viewpoint.
\newblock \emph{Acta numerica}, 30:\penalty0 87--201, 2021.

\bibitem[Beaglehole et~al.(2023)Beaglehole, Belkin, and Pandit]{beaglehole2023inconsistency}
Beaglehole, D., Belkin, M., and Pandit, P.
\newblock On the inconsistency of kernel ridgeless regression in fixed dimensions.
\newblock \emph{SIAM Journal on Mathematics of Data Science}, 5\penalty0 (4):\penalty0 854--872, 2023.

\bibitem[Beaglehole et~al.(2024)Beaglehole, S{\'u}ken{\'\i}k, Mondelli, and Belkin]{beaglehole2024average}
Beaglehole, D., S{\'u}ken{\'\i}k, P., Mondelli, M., and Belkin, M.
\newblock Average gradient outer product as a mechanism for deep neural collapse.
\newblock \emph{Advances in Neural Information Processing Systems}, 2024.

\bibitem[Belkin et~al.(2018)Belkin, Ma, and Mandal]{belkin2018understand}
Belkin, M., Ma, S., and Mandal, S.
\newblock To understand deep learning we need to understand kernel learning.
\newblock In \emph{International Conference on Machine Learning}, pp.\  541--549. PMLR, 2018.

\bibitem[Bojarski et~al.(2016)Bojarski, Choromańska, Choromanski, Fagan, Gouy-Pailler, Morvan, Sakr, Sarl{\'o}s, and Atif]{Bojarski2016StructuredAA}
Bojarski, M., Choromańska, A., Choromanski, K., Fagan, F., Gouy-Pailler, C., Morvan, A., Sakr, N., Sarl{\'o}s, T., and Atif, J.
\newblock Structured adaptive and random spinners for fast machine learning computations.
\newblock In \emph{International Conference on Artificial Intelligence and Statistics}, 2016.
\newblock URL \url{https://api.semanticscholar.org/CorpusID:7434091}.

\bibitem[Chen \& Xu(2020)Chen and Xu]{chen2020deep}
Chen, L. and Xu, S.
\newblock Deep neural tangent kernel and laplace kernel have the same rkhs.
\newblock \emph{International Conference on Representation Learning}, 2020.

\bibitem[Choromanski \& Sindhwani(2016)Choromanski and Sindhwani]{P_model}
Choromanski, K. and Sindhwani, V.
\newblock Recycling randomness with structure for sublinear time kernel expansions.
\newblock In Balcan, M.~F. and Weinberger, K.~Q. (eds.), \emph{Proceedings of The 33rd International Conference on Machine Learning}, volume~48 of \emph{Proceedings of Machine Learning Research}, pp.\  2502--2510, New York, New York, USA, 20--22 Jun 2016. PMLR.
\newblock URL \url{https://proceedings.mlr.press/v48/choromanski16.html}.

\bibitem[Choromanski et~al.(2017)Choromanski, Rowland, and Weller]{ROM_}
Choromanski, K., Rowland, M., and Weller, A.
\newblock The unreasonable effectiveness of structured random orthogonal embeddings.
\newblock In \emph{Proceedings of the 31st International Conference on Neural Information Processing Systems}, NIPS'17, pp.\  218–227, Red Hook, NY, USA, 2017. Curran Associates Inc.
\newblock ISBN 9781510860964.

\bibitem[Choromanski et~al.(2018)Choromanski, Rowland, Sarlos, Sindhwani, Turner, and Weller]{geom_random}
Choromanski, K., Rowland, M., Sarlos, T., Sindhwani, V., Turner, R., and Weller, A.
\newblock The geometry of random features.
\newblock In Storkey, A. and Perez-Cruz, F. (eds.), \emph{Proceedings of the Twenty-First International Conference on Artificial Intelligence and Statistics}, volume~84 of \emph{Proceedings of Machine Learning Research}, pp.\  1--9. PMLR, 09--11 Apr 2018.
\newblock URL \url{https://proceedings.mlr.press/v84/choromanski18a.html}.

\bibitem[Dao et~al.(2017)Dao, Sa, and R\'{e}]{Gauss_quad}
Dao, T., Sa, C.~D., and R\'{e}, C.
\newblock Gaussian quadrature for kernel features.
\newblock In \emph{Proceedings of the 31st International Conference on Neural Information Processing Systems}, NIPS'17, pp.\  6109–6119, Red Hook, NY, USA, 2017. Curran Associates Inc.
\newblock ISBN 9781510860964.

\bibitem[Demni \& Kadri(2024)Demni and Kadri]{demni2024orthogonal}
Demni, N. and Kadri, H.
\newblock Orthogonal random features: Explicit forms and sharp inequalities.
\newblock \emph{Transactions on Machine Learning Research}, 2024.
\newblock ISSN 2835-8856.
\newblock URL \url{https://openreview.net/forum?id=FMtRZ4xzSi}.

\bibitem[Feng et~al.(2015)Feng, Hu, and Liao]{scrf}
Feng, C., Hu, Q., and Liao, S.
\newblock Random feature mapping with signed circulant matrix projection.
\newblock In \emph{Proceedings of the 24th International Conference on Artificial Intelligence}, IJCAI'15, pp.\  3490–3496. AAAI Press, 2015.
\newblock ISBN 9781577357384.

\bibitem[Gaunt(2020)]{characteristic_t}
Gaunt, R.
\newblock A simple proof of the characteristic function of student's t-distribution.
\newblock \emph{Communications in Statistics: Theory and Methods}, 2020.
\newblock ISSN 0361-0926.
\newblock \doi{10.1080/03610926.2019.1702695}.

\bibitem[Geifman et~al.(2020)Geifman, Yadav, Kasten, Galun, Jacobs, and Ronen]{geifman2020similarity}
Geifman, A., Yadav, A., Kasten, Y., Galun, M., Jacobs, D., and Ronen, B.
\newblock On the similarity between the laplace and neural tangent kernels.
\newblock \emph{Advances in Neural Information Processing Systems}, 33:\penalty0 1451--1461, 2020.

\bibitem[Ghorbani et~al.(2021)Ghorbani, Mei, Misiakiewicz, and Montanari]{ghorbani2021linearized}
Ghorbani, B., Mei, S., Misiakiewicz, T., and Montanari, A.
\newblock Linearized two-layers neural networks in high dimension.
\newblock \emph{Annals of Statistics}, 2021.

\bibitem[Ghosh \& Belkin(2023)Ghosh and Belkin]{ghosh2023universal}
Ghosh, N. and Belkin, M.
\newblock A universal trade-off between the model size, test loss, and training loss of linear predictors.
\newblock \emph{SIAM Journal on Mathematics of Data Science}, 5\penalty0 (4):\penalty0 977--1004, 2023.

\bibitem[Hui et~al.(2018)Hui, Ma, and Belkin]{Hui2018KernelMB}
Hui, L., Ma, S., and Belkin, M.
\newblock Kernel machines beat deep neural networks on mask-based single-channel speech enhancement.
\newblock \emph{ArXiv}, abs/1811.02095, 2018.
\newblock URL \url{https://api.semanticscholar.org/CorpusID:53249357}.

\bibitem[J.~M.~Chambers \& Stuck(1976)J.~M.~Chambers and Stuck]{chambers1976method}
J.~M.~Chambers, C. L.~M. and Stuck, B.~W.
\newblock A method for simulating stable random variables.
\newblock \emph{Journal of the American Statistical Association}, 71\penalty0 (354):\penalty0 340--344, 1976.
\newblock \doi{10.1080/01621459.1976.10480344}.
\newblock URL \url{https://www.tandfonline.com/doi/abs/10.1080/01621459.1976.10480344}.

\bibitem[Jacot et~al.(2018)Jacot, Gabriel, and Hongler]{jacot2018neural}
Jacot, A., Gabriel, F., and Hongler, C.
\newblock Neural tangent kernel: Convergence and generalization in neural networks.
\newblock \emph{Advances in neural information processing systems}, 31, 2018.

\bibitem[Joarder \& Ali(1996)Joarder and Ali]{joarder1996characteristic}
Joarder, A.~H. and Ali, M.~M.
\newblock On the characteristic function of the multivariate t-distribution.
\newblock \emph{Pakistan Journal of Statistics}, 12:\penalty0 55--62, 1996.

\bibitem[Kar \& Karnick(2012)Kar and Karnick]{pmlr-v22-kar12}
Kar, P. and Karnick, H.
\newblock Random feature maps for dot product kernels.
\newblock In Lawrence, N.~D. and Girolami, M. (eds.), \emph{Proceedings of the Fifteenth International Conference on Artificial Intelligence and Statistics}, volume~22 of \emph{Proceedings of Machine Learning Research}, pp.\  583--591, La Palma, Canary Islands, 21--23 Apr 2012. PMLR.
\newblock URL \url{https://proceedings.mlr.press/v22/kar12.html}.

\bibitem[Karoui(2013)]{karoui2013asymptotic}
Karoui, N.~E.
\newblock Asymptotic behavior of unregularized and ridge-regularized high-dimensional robust regression estimators: rigorous results.
\newblock \emph{arXiv preprint arXiv:1311.2445}, 2013.

\bibitem[Kimeldorf \& Wahba(1971)Kimeldorf and Wahba]{kimeldorf1971some}
Kimeldorf, G. and Wahba, G.
\newblock Some results on tchebycheffian spline functions.
\newblock \emph{Journal of mathematical analysis and applications}, 33\penalty0 (1):\penalty0 82--95, 1971.

\bibitem[Kotz \& Nadarajah(2004)Kotz and Nadarajah]{Kotz_Nadarajah_2004}
Kotz, S. and Nadarajah, S.
\newblock \emph{Multivariate T-Distributions and Their Applications}.
\newblock Cambridge University Press, 2004.

\bibitem[Le et~al.(2013)Le, Sarl\'{o}s, and Smola]{Fastfood}
Le, Q., Sarl\'{o}s, T., and Smola, A.
\newblock Fastfood: approximating kernel expansions in loglinear time.
\newblock In \emph{Proceedings of the 30th International Conference on International Conference on Machine Learning - Volume 28}, ICML'13, pp.\  III–244–III–252. JMLR.org, 2013.

\bibitem[Lee et~al.(2017)Lee, Bahri, Novak, Schoenholz, Pennington, and Sohl-Dickstein]{lee2017deep}
Lee, J., Bahri, Y., Novak, R., Schoenholz, S.~S., Pennington, J., and Sohl-Dickstein, J.
\newblock Deep neural networks as gaussian processes.
\newblock \emph{arXiv preprint arXiv:1711.00165}, 2017.

\bibitem[Li(2017)]{NRFF}
Li, P.
\newblock Linearized gmm kernels and normalized random fourier features.
\newblock In \emph{Proceedings of the 23rd ACM SIGKDD International Conference on Knowledge Discovery and Data Mining}, KDD '17, pp.\  315–324, New York, NY, USA, 2017. Association for Computing Machinery.
\newblock ISBN 9781450348874.
\newblock \doi{10.1145/3097983.3098081}.
\newblock URL \url{https://doi.org/10.1145/3097983.3098081}.

\bibitem[Liu et~al.(2021)Liu, Huang, Chen, and Suykens]{liu2021random}
Liu, F., Huang, X., Chen, Y., and Suykens, J.~A.
\newblock Random features for kernel approximation: A survey on algorithms, theory, and beyond.
\newblock \emph{IEEE Transactions on Pattern Analysis and Machine Intelligence}, 44\penalty0 (10):\penalty0 7128--7148, 2021.

\bibitem[Ma \& Belkin(2019)Ma and Belkin]{ma2019kernel}
Ma, S. and Belkin, M.
\newblock Kernel machines that adapt to gpus for effective large batch training.
\newblock \emph{Proceedings of Machine Learning and Systems}, 1:\penalty0 360--373, 2019.

\bibitem[Mallinar et~al.(2022)Mallinar, Simon, Abedsoltan, Pandit, Belkin, and Nakkiran]{mallinar2022benign}
Mallinar, N., Simon, J., Abedsoltan, A., Pandit, P., Belkin, M., and Nakkiran, P.
\newblock Benign, tempered, or catastrophic: Toward a refined taxonomy of overfitting.
\newblock \emph{Advances in Neural Information Processing Systems}, 35:\penalty0 1182--1195, 2022.

\bibitem[Mallinar et~al.(2024)Mallinar, Beaglehole, Zhu, Radhakrishnan, Pandit, and Belkin]{mallinar2024emergence}
Mallinar, N., Beaglehole, D., Zhu, L., Radhakrishnan, A., Pandit, P., and Belkin, M.
\newblock Emergence in non-neural models: grokking modular arithmetic via average gradient outer product.
\newblock \emph{arXiv preprint arXiv:2407.20199}, 2024.

\bibitem[Matthews et~al.(2018)Matthews, Rowland, Hron, Turner, and Ghahramani]{matthews2018gaussian}
Matthews, A. G. d.~G., Rowland, M., Hron, J., Turner, R.~E., and Ghahramani, Z.
\newblock Gaussian process behaviour in wide deep neural networks.
\newblock \emph{arXiv preprint arXiv:1804.11271}, 2018.

\bibitem[Mei \& Montanari(2022)Mei and Montanari]{mei2022generalization}
Mei, S. and Montanari, A.
\newblock The generalization error of random features regression: Precise asymptotics and the double descent curve.
\newblock \emph{Communications on Pure and Applied Mathematics}, 75\penalty0 (4):\penalty0 667--766, 2022.

\bibitem[Munkhoeva et~al.(2018)Munkhoeva, Kapushev, Burnaev, and Oseledets]{Quad_based_feats}
Munkhoeva, M., Kapushev, Y., Burnaev, E., and Oseledets, I.
\newblock Quadrature-based features for kernel approximation.
\newblock In \emph{Proceedings of the 32nd International Conference on Neural Information Processing Systems}, NIPS'18, pp.\  9165–9174, Red Hook, NY, USA, 2018. Curran Associates Inc.

\bibitem[Muyskens et~al.(2024)Muyskens, Priest, Goumiri, and Schneider]{muyskens2024correspondence}
Muyskens, A., Priest, B.~W., Goumiri, I.~R., and Schneider, M.~D.
\newblock Correspondence of nngp kernel and the mat{\'e}rn kernel.
\newblock \emph{arXiv preprint arXiv:2410.08311}, 2024.

\bibitem[Nolan(2005)]{nolan2005multivariate}
Nolan, J.
\newblock Multivariate stable densities and distribution functions: general and elliptical case.
\newblock \emph{Deutsche Bundesbank's Annual Fall Conference}, 2005.

\bibitem[Nolan(2020)]{nolan2020univariate}
Nolan, J.
\newblock \emph{Univariate Stable Distributions: Models for Heavy Tailed Data}.
\newblock Springer International Publishing, 2020.
\newblock ISBN 9783030529154.

\bibitem[Pedregosa et~al.(2011)Pedregosa, Varoquaux, Gramfort, Michel, Thirion, Grisel, Blondel, Prettenhofer, Weiss, Dubourg, Vanderplas, Passos, Cournapeau, Brucher, Perrot, and Duchesnay]{scikit-learn}
Pedregosa, F., Varoquaux, G., Gramfort, A., Michel, V., Thirion, B., Grisel, O., Blondel, M., Prettenhofer, P., Weiss, R., Dubourg, V., Vanderplas, J., Passos, A., Cournapeau, D., Brucher, M., Perrot, M., and Duchesnay, E.
\newblock Scikit-learn: Machine learning in {P}ython.
\newblock \emph{Journal of Machine Learning Research}, 12:\penalty0 2825--2830, 2011.

\bibitem[Radhakrishnan et~al.(2024{\natexlab{a}})Radhakrishnan, Beaglehole, Pandit, and Belkin]{radhakrishnan2024mechanism}
Radhakrishnan, A., Beaglehole, D., Pandit, P., and Belkin, M.
\newblock Mechanism for feature learning in neural networks and backpropagation-free machine learning models.
\newblock \emph{Science}, 383\penalty0 (6690):\penalty0 1461--1467, 2024{\natexlab{a}}.

\bibitem[Radhakrishnan et~al.(2024{\natexlab{b}})Radhakrishnan, Cai, Weir, Moy, and Uhler]{radhakrishnan2024synthetic}
Radhakrishnan, A., Cai, C., Weir, B.~A., Moy, C., and Uhler, C.
\newblock Synthetic lethality screening with recursive feature machines.
\newblock \emph{Cancer Research}, 84\penalty0 (6\_Supplement):\penalty0 897--897, 2024{\natexlab{b}}.

\bibitem[Rahimi \& Recht(2007)Rahimi and Recht]{rahimi2007random}
Rahimi, A. and Recht, B.
\newblock Random features for large-scale kernel machines.
\newblock In Platt, J., Koller, D., Singer, Y., and Roweis, S. (eds.), \emph{Advances in Neural Information Processing Systems}, volume~20. Curran Associates, Inc., 2007.
\newblock URL \url{https://proceedings.neurips.cc/paper_files/paper/2007/file/013a006f03dbc5392effeb8f18fda755-Paper.pdf}.

\bibitem[Rakhlin \& Zhai(2019)Rakhlin and Zhai]{rakhlin2019consistency}
Rakhlin, A. and Zhai, X.
\newblock Consistency of interpolation with laplace kernels is a high-dimensional phenomenon.
\newblock In \emph{Conference on Learning Theory}, pp.\  2595--2623. PMLR, 2019.

\bibitem[Rudin(2017)]{rudin2017fourier}
Rudin, W.
\newblock \emph{Fourier analysis on groups}.
\newblock Courier Dover Publications, 2017.

\bibitem[Samoradnitsky(2017)]{samoradnitsky2017stable}
Samoradnitsky, G.
\newblock \emph{Stable non-Gaussian random processes: stochastic models with infinite variance}.
\newblock Routledge, 2017.

\bibitem[Sch{\"o}lkopf et~al.(2001)Sch{\"o}lkopf, Herbrich, and Smola]{scholkopf2001generalized}
Sch{\"o}lkopf, B., Herbrich, R., and Smola, A.~J.
\newblock A generalized representer theorem.
\newblock In \emph{International conference on computational learning theory}, pp.\  416--426. Springer, 2001.

\bibitem[Simon et~al.(2024)Simon, Karkada, Ghosh, and Belkin]{simon2024more}
Simon, J.~B., Karkada, D., Ghosh, N., and Belkin, M.
\newblock More is better: when infinite overparameterization is optimal and overfitting is obligatory.
\newblock In \emph{The Twelfth International Conference on Learning Representations}, 2024.

\bibitem[Stein \& Weiss(1971)Stein and Weiss]{Stein1971}
Stein, E.~M. and Weiss, G.
\newblock \emph{Introduction to Fourier Analysis on Euclidean Spaces (PMS-32)}.
\newblock Princeton University Press, 1971.
\newblock ISBN 9780691080789.
\newblock URL \url{http://www.jstor.org/stable/j.ctt1bpm9w6}.

\bibitem[Steinwart \& Christmann(2008)Steinwart and Christmann]{steinwart2008support}
Steinwart, I. and Christmann, A.
\newblock \emph{Support Vector Machines}.
\newblock Information Science and Statistics. Springer New York, 2008.
\newblock ISBN 9780387772424.
\newblock URL \url{https://books.google.co.in/books?id=HUnqnrpYt4IC}.

\bibitem[Sutherland \& Schneider(2015)Sutherland and Schneider]{sutherland2015error}
Sutherland, D.~J. and Schneider, J.
\newblock On the error of random fourier features.
\newblock \emph{Proceedings of the Thirty-First Conference on Uncertainty in Artificial Intelligence}, 2015.

\bibitem[Sutradhar(1986)]{sutradhar1986characteristic}
Sutradhar, B.~C.
\newblock On the characteristic function of multivariate student t-distribution.
\newblock \emph{The Canadian Journal of Statistics/La Revue Canadienne de Statistique}, pp.\  329--337, 1986.

\bibitem[Wacker et~al.(2024)Wacker, Kanagawa, and Filippone]{wacker2024improved}
Wacker, J., Kanagawa, M., and Filippone, M.
\newblock Improved random features for dot product kernels.
\newblock \emph{Journal of Machine Learning Research}, 25\penalty0 (235):\penalty0 1--75, 2024.

\bibitem[Weron(1996)]{weron1996chambers}
Weron, R.
\newblock On the chambers-mallows-stuck method for simulating skewed stable random variables.
\newblock \emph{Statistics \& Probability Letters}, 28\penalty0 (2):\penalty0 165--171, 1996.
\newblock ISSN 0167-7152.
\newblock \doi{https://doi.org/10.1016/0167-7152(95)00113-1}.
\newblock URL \url{https://www.sciencedirect.com/science/article/pii/0167715295001131}.

\bibitem[Williams \& Rasmussen(2006)Williams and Rasmussen]{williams2006gaussian}
Williams, C.~K. and Rasmussen, C.~E.
\newblock \emph{Gaussian processes for machine learning}, volume~2.
\newblock MIT press Cambridge, MA, 2006.

\bibitem[Yu et~al.(2016)Yu, Suresh, Choromanski, Holtmann-Rice, and Kumar]{yu2016orthogonal}
Yu, F. X.~X., Suresh, A.~T., Choromanski, K.~M., Holtmann-Rice, D.~N., and Kumar, S.
\newblock Orthogonal random features.
\newblock \emph{Advances in neural information processing systems}, 29, 2016.

\end{thebibliography}
\bibliographystyle{icml2025}

\onecolumn
\appendix
{\center\bf\huge Appendices}
\section{Proofs and derivations}
\label{appendix:proofs}
The following proof is provided in \cite{Stein1971}.
\begin{proof}[Proof of \Cref{lemma:fourier_laplace}]
    Note the following fact (derived at the end),
    \begin{align}e^{-\beta}=  \frac{1}{\sqrt{\pi}} \int_0^{\infty} \frac{\e^{-
    u}}{\sqrt{u}} \e^{- \frac{\beta^2}{4 u} } \dif u\qquad \forall\, \beta\in\Real \label{eq:exp_as_integral}
    \end{align}
    Observe that
    \begin{align*}
    &\int_{\Real^d} e^{-\norm{\x}} e^{- j\w\tran\x}\mathop{d\x} \\
    &=\int_{\Real^d}
    \left[ \frac{1}{\sqrt{\pi}} \int_0^{\infty} \frac{e^{- u}}{\sqrt{u}}
    e^{- \frac{\norm{\x}^2}{4u} }\mathop{du} \right] e^{- j
    \w\tran\x}\mathop{d\x} \\
    &= \frac{1}{\sqrt{\pi}} \int_0^{\infty} \frac{e^{- u}}{\sqrt{u}}
    \left[ \int_{\Real^d} e^{- \frac{\norm{\x}^2}{4 u} }
    e^{- j \w\tran \x} \mathop{d\x} \right] \mathop{du}\\
    &= \frac{1}{\sqrt{\pi}} \int_0^{\infty} \frac{e^{- u}}{\sqrt{u}}
    \left[ \left( \sqrt{4 \pi u} \right)^d e^{-\norm{\w} ^2 u}
    \right] \mathop{du}  \\
    &= \frac{2^d \sqrt{\pi^d}}{\sqrt{\pi}}  \int_0^{\infty} 
    u^{\left( \frac{d - 1}{2} \right)} e^{-(1+\norm{\w}^2) u}
    \mathop{du} \\
    &= 2^d \sqrt{\pi^{d - 1}} \Gamma \left( \frac{d + 1}{2} \right)  (1 +
    \norm{\w}^2)^{- \left( \frac{d + 1}{2} \right)}
    \end{align*}

    Finally, we prove \cref{eq:exp_as_integral}. For this we state equalities
    \begin{align*}
        \e^{-\beta} &= \frac{2}{\pi}\int_0^\infty\frac{\cos(\beta x)}{1+x^2}\dif x \\
        \frac{1}{1+x^2} &= \int_0^\infty\e^{-(1+x^2)u}\dif u
    \end{align*}
    Some quick manipulations lead us to the desired result:
    \begin{align*}
        \e^{-\beta} &= \frac{2}{\pi}\int_0^\infty\frac{\cos(\beta x)}{1+x^2}\dif x \\
        &= \frac{2}{\pi}\int_0^\infty \cos(\beta x) \left\{ \int_0^\infty\e^{-(1+x^2)u}\dif u \right\}\dif x\\
        &= \frac{2}{\pi}\int_0^\infty \e^{-u}\left\{ \int_0^\infty\e^{-ux^2}\cos(\beta x)\dif x \right\}\dif u \\
        &= \frac{1}{\sqrt{\pi}} \int_0^{\infty} \frac{\e^{-u}}{\sqrt{u}} e^{- \frac{\beta^2}{4 u} } \dif u
    \end{align*}

The case of $M\neq \sigma^2 I_d$ is straightforward from the duality of the norms in the Fourier transform.
\end{proof}

\begin{proof}[Proof of \Cref{lemma:sample_norm_of_t}]
        Changing variables to $(r, \theta)$. Let $r = \norm w$ and $B(z_1, z_2) = \frac{\Gamma(z_1)\Gamma(z_2)}{\Gamma(z_1+z_2)}$ denote the Beta function.
    \begin{align*}
\int_{\Real^d}{\frac{\Gamma(\nu+\frac{d}2)}{\Gamma(\nu) \sqrt{\round{2\nu\pi\sigma^2}^{d}}}}
     \left[1+\frac{\norm{w}^2}{2\sigma^2 \nu}\right]^{-(\nu+\frac{d}2)} \dif {w} &=1\\
 \int_{\Real}\frac{\Gamma(\nu+\frac{d}2)r^{d-1}}{\Gamma(\nu) \sqrt{\round{2\nu\pi\sigma^2}^{d}}}\frac{2 \pi^{d/2}}{\Gamma(\frac d2)}
         \left[1+\frac{r^2}{2\sigma^2 \nu}\right]^{-(\nu+\frac{d}2)} \dif r &=1 
    \end{align*}
We note that the integrand itself is the radial measure we want. We set $B \left( \frac{d}{2}, \nu \right) = \tfrac{\Gamma \left( \frac{d}{2}\right) \Gamma \left(\nu \right)}{\Gamma\left(\frac{d+2\nu}{2}\right)}$
    \begin{align*}
        f(r) &= \frac{2r^{d-1}}{\sigma^d\round{2\nu}^{d/2}B(\frac{d}{2}, \nu)\left(1+\frac{r^2}{2\sigma^2 \nu}\right)^{\frac{d+2\nu}{2}}}\\
       f(r) &= \frac{2\left(\frac{r}{\sigma\sqrt{2\nu}}\right)^{d-1}}{\sigma\sqrt{2\nu}B(\frac{d}{2}, \nu)\left(1+\frac{r^2}{2\sigma^2\nu}\right)^{\frac{d+2\nu}{2}}}
    \end{align*}
Note that the density function of the $\textsf{GBP}$ is given by  
\begin{align*}
    f(x;\alpha, \beta, p, q) = \frac{p\left(\frac{x}{q}\right)^{\alpha p-1}}{qB(\alpha, \beta)\left(1+\left(\frac{x}{q}\right)^p\right)^{\alpha + \beta}}
\end{align*}
Which exactly matches for parameter values $\alpha=\frac{d}{2}$, $\beta = \nu, p = 2, q = \sigma\sqrt{2\nu}$.
\end{proof}

\section{Other experimental details}
\label{appendix:expts}
For the more general location family stable distributions with location parameter $\mu\in\Real$ would output with an offset of $\mu.$
\begin{algorithm}[H]
\caption{\label{algo:cms}Chambers Mallows Stuck algorithm \cite{chambers1976method}.}
\begin{algorithmic}
\REQUIRE Random samples $V \sim \text{Uniform}(-\frac\pi2, \frac\pi2), $ and $W \sim \text{Exponential}(1)$,\\ shape parameters $\alpha \in (0,2],$ and $ \beta \in [-1, 1],$ and scale parameter $\sigma>0.$
\STATE {\bf Output:} Sample from univariate $\alpha$-stable distribution $S(\alpha, \beta, \sigma)$
\STATE Set $B_{\alpha, \beta} = \frac{1}{\alpha}\arctan(\beta \tan(\frac{\pi \alpha}{2}))$
\STATE Set $C_{\alpha, \beta} = (1+\beta^2\tan^2(\pi\alpha/2))^\frac{1}{2\alpha} $
\RETURN Set $\sigma\cdot C_{\alpha, \beta} \frac{\sin(\alpha(V + B_{\alpha, \beta}))}{(\cos(V))^{1/\alpha}}\left(\frac{\cos(V - \alpha(V + B_{\alpha, \beta}))}{W}\right)^{\frac{1-\alpha}{\alpha}}$
\end{algorithmic}
\end{algorithm}

\begin{figure}
\includegraphics[width=0.95\textwidth]{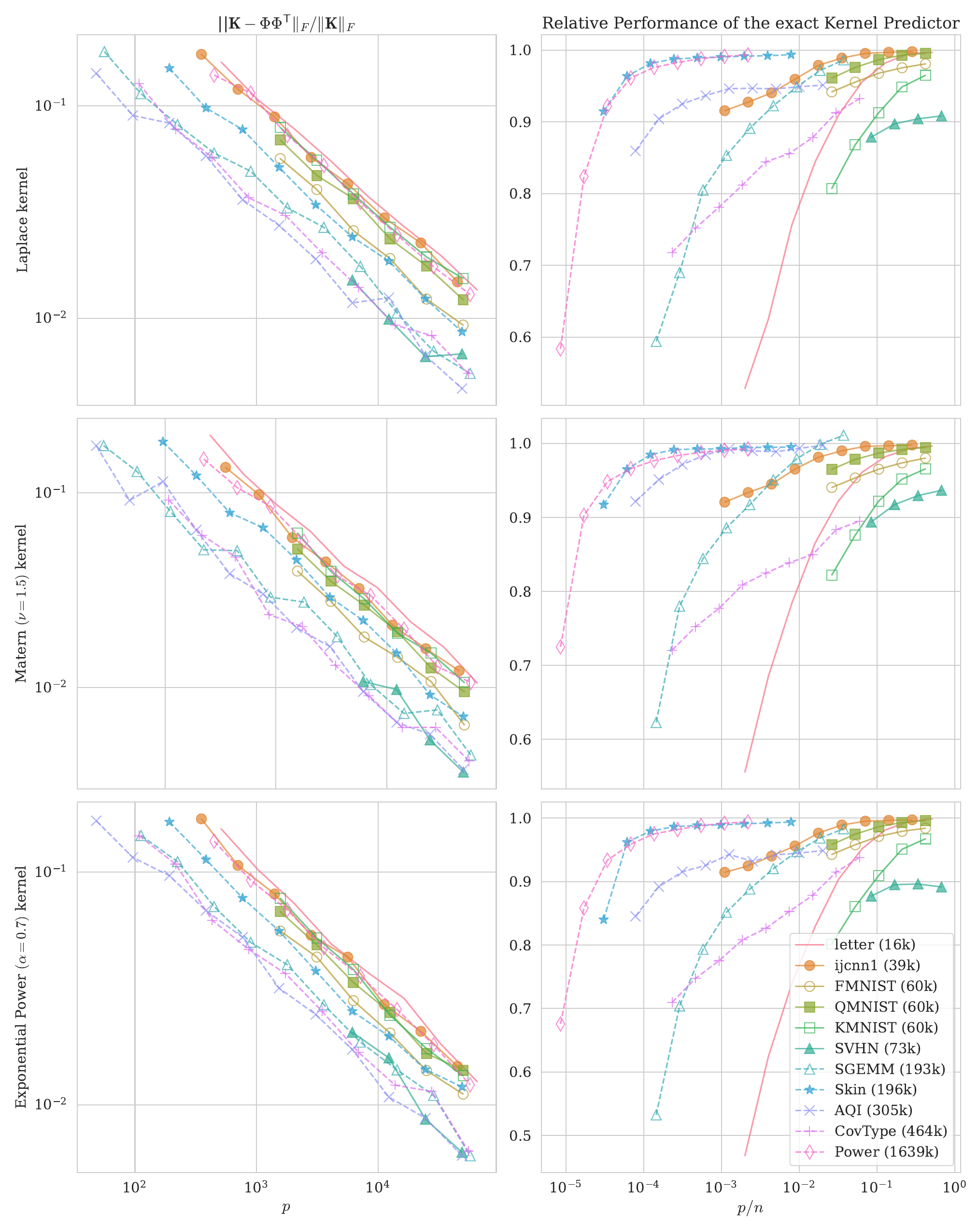}
\caption{\label{fig:RFF_eval}Evaluation of RFF. \textbf{(Left)}: relative approximation error for the kernel matrix various datasets in frobenius norm, for $n=10,000$ samples. Error measured in other norms (operator, nuclear) are mentioned in Figures \ref{fig:approx_op_norm}, \ref{fig:approx_nuc_norm}. \textbf{(Right)}: performance of random features predictor in comparison to exact solution via EigenPro2 \cite{ma2019kernel}.}
\end{figure}

\begin{table}[h]
\centering

\caption{\label{tab:datasets}Summary of datasets used.}
\begin{tabular}{lcccl}
\toprule
    Dataset & $d$ & $n$ & Metric & Preprocessing\\
\midrule
    letter & 16 & 20k & accuracy & Subtract mean, set $\norm x_2=1$\\ 
    icnn1 & 22 & 40k & accuracy & Set $\norm x_2=1$\\ 
    FMNIST & 784 & 60k & accuracy& Set $\norm x_2=1$\\ 
    QMNIST & 784 & 60k & accuracy & Set $\norm x_2=1$\\ 
    KMNIST & 784 & 60k & accuracy & Set $\norm x_2=1$\\ 
    SVHN & 3072 & 73k & accuracy & Set $\norm x_2=1$\\ 
    SGEMM & 14 & 193k & $R^2$ & $y\mapsto\log(y)$. Set $\norm x_2=1$  \\ 
    Skin & 3 & 196k & accuracy & Subtract mean, set $\norm x_2=1$\\ 
    AQI & 12 & 305k & $R^2$ & set $\norm x_2=1$\\ 
    CovType & 54 & 464k & $R^2$ & set $\norm x_2=1$\\ 
    Power & 7 & 1639k & $R^2$ & Standard scaler, set $\norm x_2=1$\\
\bottomrule
\end{tabular}
\end{table}

\subsection{Time comparison for kernel computation}
We present amount of speedup for kernel matrix computation compared to the \texttt{sklearn} implementation for 2 datasets for different values of $\nu$

\begin{figure}[!htb]
\begin{center}
\centerline{\includegraphics[width=\columnwidth]{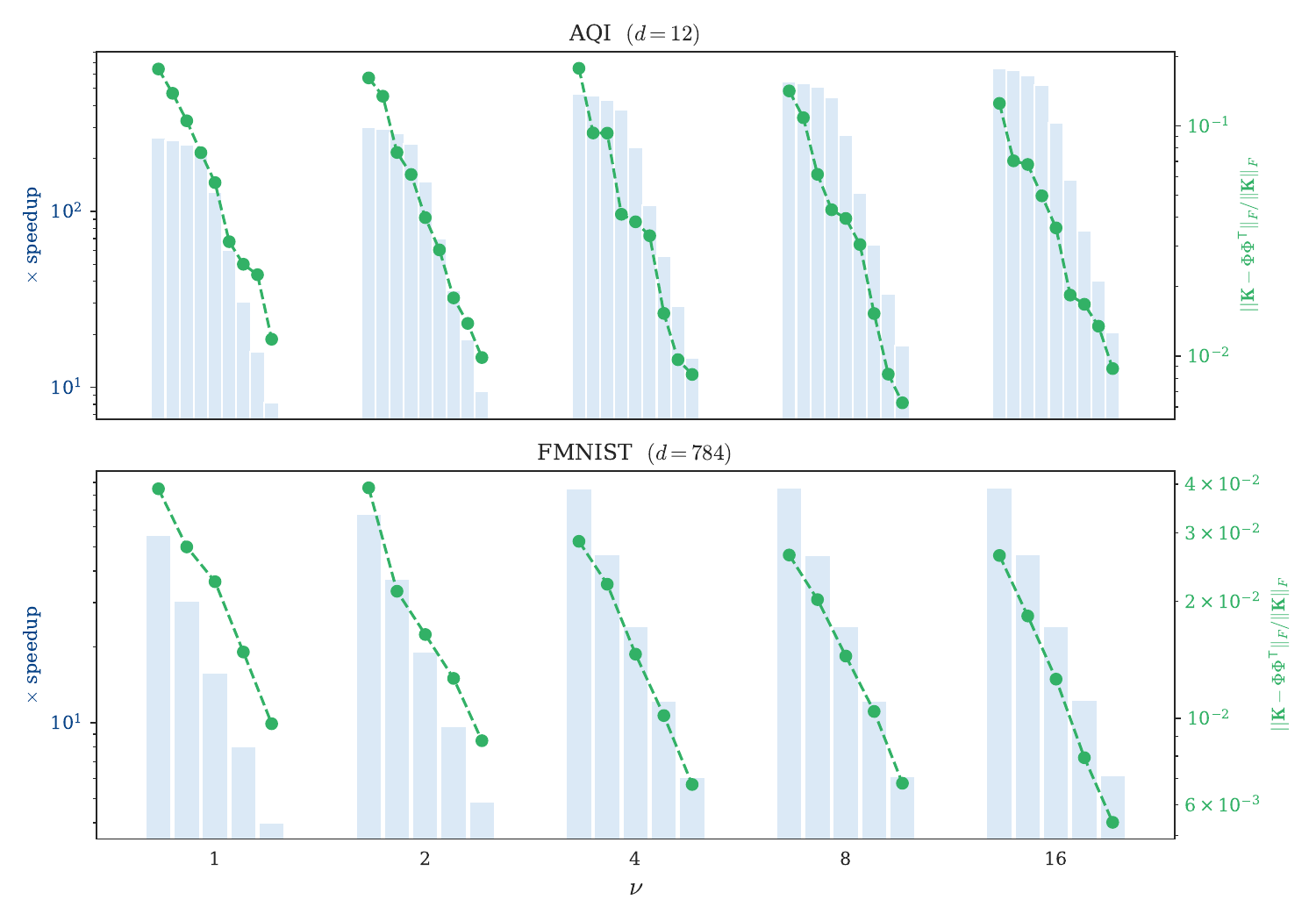}}
\caption{Amount of speedup for Kernel computation, along with relative Frobenius error for $n=10,000$ samples. \textbf{(1)}: for the AQI dataset $(d=12)$, (\textbf{2}): for the FMNIST dataset $(d=784)$.}
\end{center}
\vskip -0.2in
\end{figure}

\begin{figure}
\includegraphics[width=\textwidth]{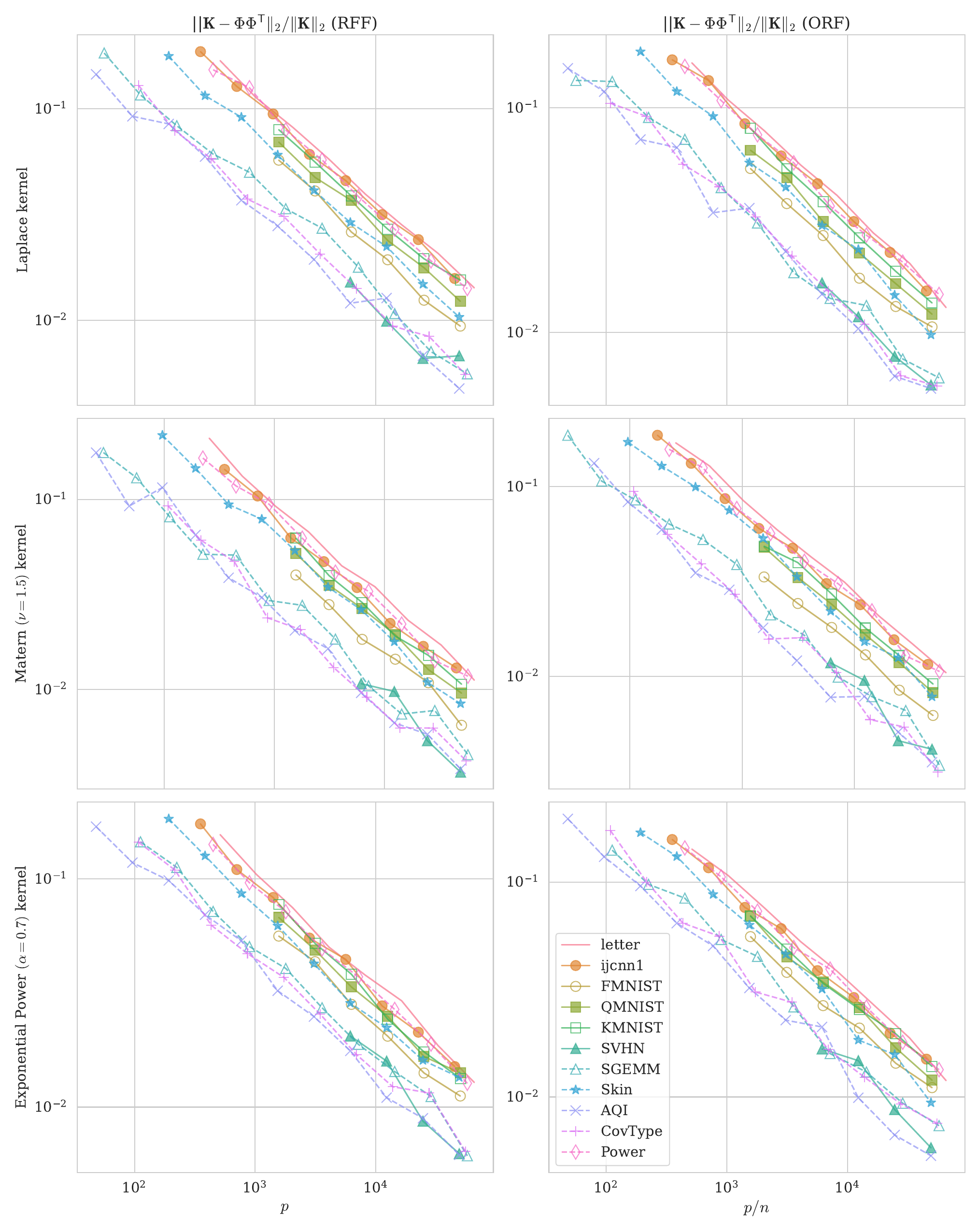}
\caption{\label{fig:approx_op_norm}Relative approximation error (operator norm) for the kernel matrix using the RFF sampling, for $n=10,000$ samples. \textbf{(Left)}: $\Phi$ is computed using the RFF sampling. \textbf{(Right)}: $\Phi$ is computed using the ORF sampling.}
\end{figure}

\begin{figure}
\includegraphics[width=\textwidth]{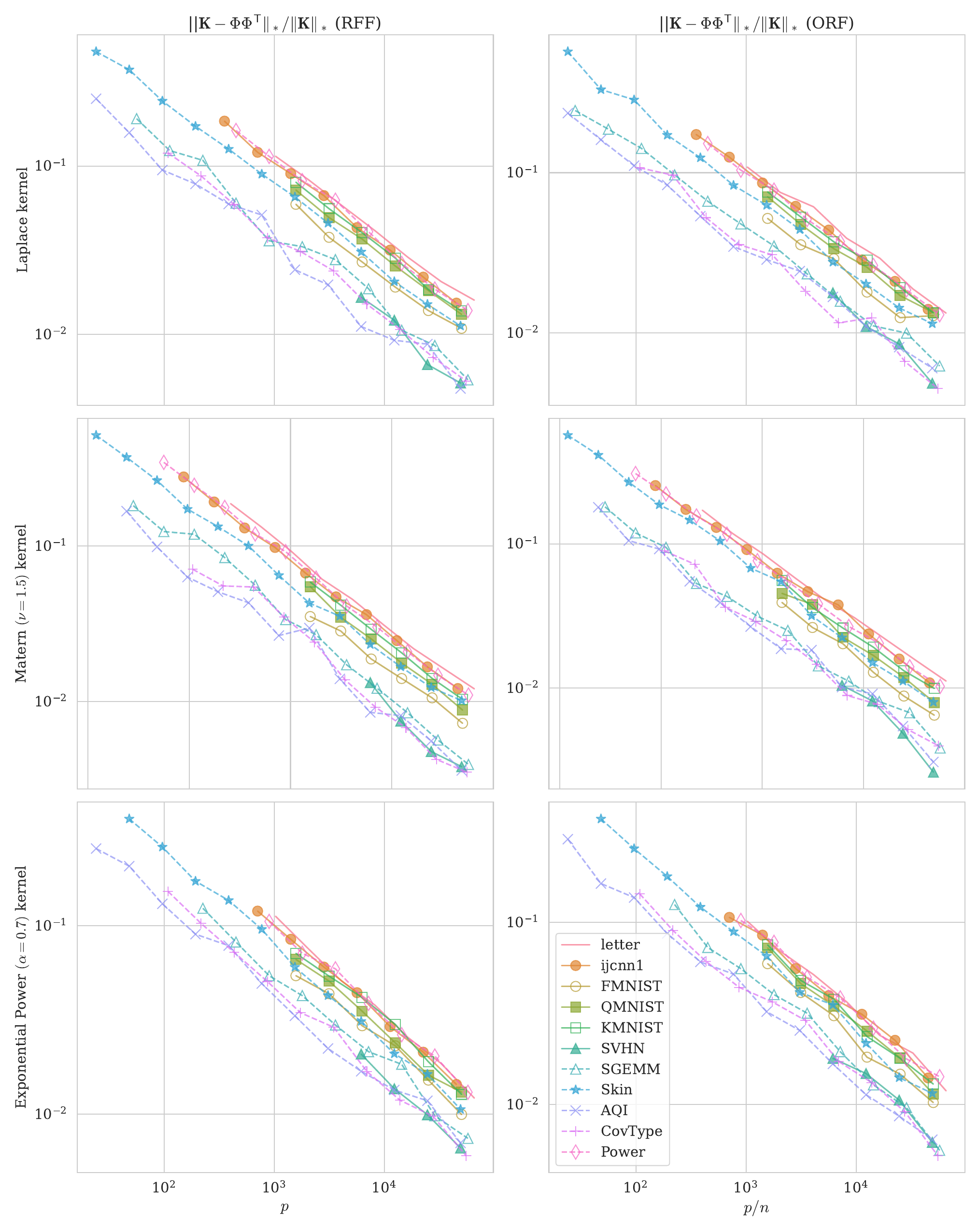}
\caption{\label{fig:approx_nuc_norm}Relative approximation error (nuclear norm) for the kernel matrix using the RFF sampling, for $n=1,000$ samples. \textbf{(Left)}: $\Phi$ is computed using the ORF sampling. \textbf{(Right)}: $\Phi$ is computed using the RFF sampling.}
\end{figure}

\begin{table}[!htb]
    \centering
    \caption{\label{tab:krr}Performance of random features predictor in comparison to exact solution via EigenPro2 \cite{ma2019kernel}. Avg Metric, Std deviation is reported over 5 experiments. Refer to \cref{tab:datasets} for Metric and pre-processing details. (Both Avg Metric and Std deviation have been multiplied by 100 below)}
    \begin{tabular}{lllcccccc}
    \toprule
        \textbf{Dataset} & Kernel& KRR & \multicolumn{3}{c}{RFF} &  \multicolumn{3}{c}{ORF}  \\
        \cmidrule(lr){4-6}  \cmidrule(lr){7-9}
        ~ & ~ & ~ & $p^{*}$ & Avg Metric & Std dev & $p^{*}$ & Avg Metric & Std dev \\ \midrule
        \textbf{letter} & Laplacian & 97.5 &  8,192  & 97.2 & 2.23E-01 &  8,192  & 97.4 & 3.32E-02 \\ 
        $d=16$ & ExpPower & 97.3 &  8,192  & 97.2 & 1.09E-01 &  8,192  & 96.9 & 5.70E-02 \\ 
        $n=20k$ & \matern & 97.9 &  8,192  & 97.5 & 8.75E-02 &  8,192  & 97.5 & 6.21E-02 \\ \midrule
        \textbf{icnn1} & Laplacian & 98.9 &  11,264  & 98.7 & 5.69E-02 &  11,264  & 98.7 & 8.92E-02 \\ 
        $d=22$ & ExpPower & 98.9 &  11,264  & 98.6 & 7.00E-02 &  11,264  & 98.7 & 6.28E-02 \\ 
        $n=40k$ & \matern & 98.9 &  11,264  & 98.7 & 7.13E-02 &  11,264  & 98.8 & 7.47E-02 \\ \midrule
        \textbf{FMNIST} & Laplacian & 90.8 &  25,088  & 89.1 & 1.16E-01 &  25,088  & 89.1 & 1.97E-01 \\ 
        $d=784$ & ExpPower & 90.3 &  25,088  & 88.8 & 1.37E-01 &  25,088  & 88.8 & 1.67E-01 \\ 
        $n=60k$ & \matern & 91.3 &  25,088  & 89.5 & 1.77E-01 &  25,088  & 89.5 & 9.85E-02 \\ \midrule
        \textbf{QMNIST} & Laplacian & 98.3 &  25,088  & 97.9 & 2.23E-02 &  25,088  & 97.9 & 3.94E-02 \\ 
        $d=784$ & ExpPower & 98.1 &  25,088  & 97.8 & 1.47E-02 &  25,088  & 97.7 & 3.16E-02 \\ 
        $n=60k$ & \matern & 98.6 &  25,088  & 98.1 & 2.69E-02 &  25,088  & 98.1 & 4.54E-02 \\ \midrule
        \textbf{KMNIST} & Laplacian & 93 &  25,088  & 89.7 & 1.54E-01 &  25,088  & 89.9 & 1.93E-01 \\ 
        $d=784$ & ExpPower & 91.9 &  25,088  & 88.9 & 2.06E-01 &  25,088  & 89.2 & 2.29E-01 \\ 
        $n=60k$ & \matern & 94.1 &  25,088  & 90.9 & 1.36E-01 &  25,088  & 90.9 & 1.54E-01 \\ \midrule
        \textbf{SVHN} & Laplacian & 80.8 &  49,152  & 73.4 & 1.82E-01 &  49,152  & 73.4 & 1.74E-01 \\ 
        $d=3072$ & ExpPower & 79.9 &  24,576  & 71.6 & 2.72E-01 &  24,576  & 71.7 & 3.11E-01 \\ 
        $n=73k$ & \matern & 80.5 &  49,152  & 75.4 & 2.02E-01 &  49,152  & 75.2 & 1.05E-01 \\ \midrule
        \textbf{SGEMM} & Laplacian & 98.4 &  7,168  & 97.1 & 6.49E-02 &  7,168  & 97.1 & 2.34E-02 \\ 
        $d=14$ & ExpPower & 98.7 &  7,168  & 97 & 2.56E-02 &  7,168  & 97.1 & 9.55E-02 \\ 
        $n=193k$ & \matern & 95.4 &  7,168  & 96.4 & 5.22E-02 &  7,168  & 96.3 & 1.97E-02 \\ \midrule
        \textbf{Skin} & Laplacian & 99.3 &  1,536  & 98.7 & 1.43E-02 &  1,536  & 98.7 & 8.40E-03 \\ 
        $d=3$ & ExpPower & 99.4 &  1,536  & 98.8 & 4.33E-02 &  1,536  & 98.7 & 3.30E-02 \\ 
        $n=196k$ & \matern & 99 &  1,536  & 98.5 & 1.32E-02 &  1,536  & 98.5 & 7.46E-03 \\ \midrule
        \textbf{AQI} & Laplacian & 81.7 &  6,144  & 77.7 & 8.15E-02 &  6,144  & 78 & 4.20E-01 \\ 
        $d=12$ & ExpPower & 82.4 &  6,144  & 78.1 & 3.11E-01 &  6,144  & 78.3 & 1.99E-01 \\ 
        $n=305k$ & \matern & 78 &  6,144  & 77.8 & 5.52E-01 &  6,144  & 78 & 1.41E-01 \\ \midrule
        \textbf{CovType} & Laplacian & 95.5 &  27,648  & 89 & 3.74E-02 &  27,648  & 89 & 8.33E-02 \\ 
        $d=54$ & ExpPower & 95.8 &  27,648  & 89.9 & 8.45E-02 &  27,648  & 89.9 & 6.22E-02 \\ 
        $n=464k$ & \matern & 96.1 &  27,648  & 86 & 6.01E-02 &  27,648  & 86 & 1.97E-02 \\ \midrule
        \textbf{Power} & Laplacian & 99.6 &  3,584  & 98.9 & 9.20E-03 &  3,584  & 99 & 1.23E-02 \\ 
        $d=7$ & ExpPower & 99.5 &  3,584  & 98.9 & 7.21E-03 &  3,584  & 98.9 & 6.85E-03 \\ 
        $n=1639k$ & \matern & 99.7 &  3,584  & 99 & 1.64E-02 &  3,584  & 99 & 1.35E-02 \\ 
    \bottomrule
    \end{tabular}
\end{table}

\end{document}